\documentclass[journal,twoside,web]{ieeecolor}
\usepackage{jsen}
\usepackage{amsmath,mathtools,amssymb,amsfonts,bbm}
\let\labelindent\relax

\usepackage{cite,cleveref}
\usepackage{algorithm,algorithmic}
\usepackage{enumitem}
\usepackage{support-caption}
\usepackage{subcaption}
\usepackage{tikz}
\usetikzlibrary{shapes,arrows,calc,decorations,intersections}
\usepackage{graphicx,xcolor}
\graphicspath{{Figures/}}
\usepackage{textcomp}
\usepackage{wrapfig}

\newcommand{\upi}{\pi_{\mathrm{upper}}}
\newcommand{\vecpi}{\mathbf{\pi}}
\newcommand{\KL}[2]{\mathrm{KL}\lb #1 \Vert #2\rb}
\crefname{figure}{fig.}{figs.}
\newcommand{\hs}[1]{\hspace{#1cm}}

\definecolor{colororange}{HTML}{E65100} 
\definecolor{colorgreen}{HTML}{009688} 
\definecolor{colorblue}{HTML}{0277BB}
\definecolor{coloryellow}{HTML}{CCCC00}

\crefname{app}{Appendix}{Appendices}
\crefname{cor}{Corollary}{Corollary}
\crefname{prop}{Proposition}{Proposition}
\crefname{lemma}{Lemma}{Lemma}
\crefname{defn}{Definition}{Definition}
\crefname{conj}{Conjecture}{Conjecture}
\crefname{exam}{Example}{Example}
\crefname{supp}{Supplemental Section}{Supplemental Section}

\newtheorem{prop}{Proposition}

\newcommand{\bs}{\boldsymbol}
\newcommand{\bb}{\mathbb}
\newcommand{\mcal}{\mathcal}

\newcommand{\one}{\bs{1}}

\newcommand{\lb}{\left(}
\newcommand{\rb}{\right)}
\newcommand{\ls}{\left[}
\newcommand{\rs}{\right]}
\newcommand{\lc}{\left\{}
\newcommand{\rc}{\right\}}

\newcommand{\lv}{\left\vert}
\newcommand{\rv}{\right\vert}
\newcommand{\lV}{\left\Vert}
\newcommand{\rV}{\right\Vert}

\newcommand{\LRV}[1]{{\left\vert\kern-0.25ex\left\vert\kern-0.25ex\left\vert #1 \right\vert\kern-0.25ex\right\vert\kern-0.25ex\right\vert}}

\newcommand{\expect}[2]{\bb{E}_{#1}\lc#2\rc}

\newcommand{\nth}{^\mathsf{th}}
\newcommand{\tran}{^{\mathsf{T}}}

\newcommand{\bbP}{\bb{P}}

\newcommand{\bbR}{\bb{R}}

\newcommand{\calA}{\mcal{A}}

\newcommand{\calP}{\mcal{P}}

\newcommand{\vecd}{\bs{d}}
\newcommand{\vece}{\bs{e}}

\newcommand{\vech}{\bs{h}}

\newcommand{\vecq}{\bs{q}}

\newcommand{\vecx}{\bs{x}}
\newcommand{\vecy}{\bs{y}}

\def\BibTeX{{\rm B\kern-.05em{\sc i\kern-.025em b}\kern-.08em
    T\kern-.1667em\lower.7ex\hbox{E}\kern-.125emX}}
\definecolor{abstractbg}{rgb}{0.89804,0.94510,0.83137}
\begin{document}
\title{Anomaly Detection via Learning-Based Sequential Controlled Sensing
}
\author{Geethu~Joseph, Chen~Zhong, M.~Cenk~Gursoy, Senem~Velipasalar, and Pramod~K.~Varshney~\IEEEmembership{Life~Fellow,~IEEE}
\thanks{This work was supported in part by the  National Science Foundation under grants ENG 60064237. The material in this paper was presented in part at the IEEE International Workshop on Signal Processing Advances in Wireless Communications, May 2020, Atlanta, GA, USA, and the IEEE Global Communications Conference, December 2020, Taipei, Taiwan.}
\thanks{G. Joseph is with the faculty of Electrical Engineering, Mathematics, and Computer Science,  Delft Technical University, 2628 XE, Netherlands (email: g.joseph@tudelft.nl). }
\thanks{C. Zhong, M. C. Gursoy, S. Velipasalar, and P. K.Varshney are with the Department of Electrical and Computer Engineering, Syracuse University, New York, 13244, USA (emails:\{czhong03,mcgursoy,svelipas,varshney\}@syr.edu.)}
}

\IEEEtitleabstractindextext{%
\fcolorbox{abstractbg}{abstractbg}{%
\begin{minipage}{\textwidth}%

\begin{wrapfigure}[14]{r}{3.6in}%

\vspace{-0.3cm}
\begin{tikzpicture}
\tikzstyle{box} = [draw=none,fill = gray,line width=1pt,text=white,rectangle,  rounded corners, text centered,node distance = 0.75cm,font=\bfseries]
\tikzstyle{box1} = [draw=white,fill=colorblue!70,text=white,rectangle,  rounded corners, text centered,font=\bfseries]
\tikzstyle{cloud} = [draw=none,text=colorblue,rectangle,  rounded corners, text centered,font=\bfseries]
\tikzstyle{comment} = [midway, above, sloped]
\tikzstyle{line} = [-,line width=1pt]
\tikzstyle{linea} = [->,dashed,line width=1pt,draw=colororange]

\node[box](P1){Process 1};
\node[box,below of =P1](P2){Process 2};
\node[node distance = 0.5cm,below of =P2](Pd){\textbf{$\vdots$}};
\node[box,node distance = 0.75cm,below of =Pd](P4){Process $N$};
\node[box,node distance = 2cm, left of = Pd,rotate=90](Cor){Joint distribution};

\node[above of = P1,node distance = 0.7cm, xshift=-0.75cm]{\color{colororange}\textbf{Environment}};
\draw[colororange,line width=1pt] ($(P1.north west)+(-1.8,0.7)$)  rectangle ($(P4.south east)+(0.3,-0.6)$);
\node[node distance = 2cm, right of = Pd](SN){};

\node[node distance = 5.1cm, right of = Pd,box1, text width=2cm](FC){Our learning algorithm};
\node[node distance = 2cm, below of = FC, text width=3cm,cloud](Stop){Anomalous processes' indices};

  \draw [colororange, ->,line width=1pt] ([yshift=-0.2cm,]SN.east)--([yshift=-0.2cm,xshift=-0.5cm]FC.west) node [midway, below, text width=2.3cm] (TextNode) {Noisy sensor measurements};

\draw [colorblue, ->,line width=1pt] ([yshift=0.2cm, xshift=-0.5cm]FC.west)--([yshift=0.2cm]SN.east) node [midway, above, sloped,text width=2.3cm,text centered] (TextNode) {Dynamic sensor selection};

\draw [colorblue, ->,line width=1pt] (FC.south)--(Stop.north) node [midway,right] (TextNode) {Output};

\draw [gray, ->,line width=1pt,dotted](Cor)--(P1.west);
\draw [gray, ->,line width=1pt,dotted](Cor)--(P2.west);
\draw [gray, ->,line width=1pt,dotted](Cor)--(P4.west);
\end{tikzpicture}
\end{wrapfigure}%
\begin{abstract}
In this paper, we address the problem of detecting anomalies among a given set of binary processes via learning-based controlled sensing. Each process is parameterized by a binary random variable indicating whether the process is anomalous. To identify the anomalies, the decision-making agent is allowed to observe a subset of the processes at each time instant. Also, probing each process has an associated cost. Our objective is to design a sequential selection policy that dynamically determines which processes to observe at each time with the goal to minimize the delay in making the decision and the total sensing cost. We cast this problem as a sequential hypothesis testing problem within the framework of Markov decision processes. This formulation utilizes both a Bayesian log-likelihood ratio-based reward and an entropy-based reward. The problem is then solved using two approaches: 1) a deep reinforcement learning-based approach where we design both deep Q-learning and policy gradient actor-critic algorithms; and 2) a deep active inference-based approach. Using numerical experiments, we demonstrate the efficacy of our algorithms and show that our algorithms adapt to any unknown statistical dependence pattern of the processes.
\end{abstract}

\end{minipage}}}
\maketitle

\section{Introduction}
\emph{Sequential controlled sensing} refers to a stochastic framework in which an agent sequentially controls the process of acquiring observations. The goal here is to minimize the cost of making observations while satisfying the inference objectives. We consider a sequential controlled sensing problem in the context of anomaly detection wherein there are $N$ processes, each of which can be in either a normal or an anomalous condition. Our goal is to identify the anomalies among the given processes. To this end, the decision-making agent sequentially chooses a subset of processes (or equivalently sensors monitoring these processes) at every time instant, probes them, and obtain estimates of their conditions. The agent generally obtains noisy observations, i.e., the observed condition may get flipped from the actual condition with a certain probability.  This paradigm is encountered in many practical applications such as remote health monitoring, assembly lines, structural health monitoring and Internet of Things (IoT). In such applications, the objective is to identify the anomalies among a given set of different (not necessarily independent) functionalities of a system~\cite{chung2006remote,bujnowski2013enhanced}. Each sensor monitors a different functionality and sends observations to the agent over a communication link. The received observation may be distorted due to the unreliable nature of the sensor hardware and/or the noisy link (e.g., a wireless channel) between the sensor and the agent. Hence, the agent needs to probe each process multiple times before declaring one or more of the processes anomalous with the desired confidence. Repeatedly probing all the processes allows the agent to find any potential system malfunction or anomaly quickly, but this incurs a higher energy consumption that reduces the life span of the network. Therefore, we address the question of how the agent must sequentially choose the subset of processes to accurately detect the anomalies while minimizing the delay and cost of making observations.

We start with a brief literature review. A classical approach to solve the sequential process selection problem for anomaly detection is based on the active hypothesis testing framework~\cite{zhong2019deep,joseph2020anomaly}. Here, the decision-making agent constructs a hypothesis corresponding to each of the possible conditions of the processes and determines which one of these hypotheses is true. The goal of active hypothesis testing is to infer the true hypothesis by collecting relevant data sequentially until sufficiently strong evidence is gathered. In \cite{chernoff1959sequential}, Chernoff proposed a randomized strategy and established its asymptotic optimality. This seminal work in~\cite{chernoff1959sequential} was followed by several other studies that investigated active hypothesis testing under different settings~\cite{bessler1960theory,naghshvar2012extrinsic,naghshvar2013active,
franceschetti2016chernoff,
huang2018active}. These studies investigated the theoretical aspects of the problem and presented a few model-based algorithms to solve the problem. However, these model-based algorithms are designed under simplified modeling assumptions. This has motivated the researchers to design data-driven deep learning algorithms for active hypothesis testing~\cite{kartik2018policy,zhong2019deep,joseph2020anomaly,
joseph2020anomaly2,joseph2021temporal,joseph2021scalable,joseph2021scalable2}. These algorithms are not only model free and thus, more flexible than the traditional algorithms, but they also possess reduced computational complexity. It should be noted that the classical sequential hypothesis testing framework does not incorporate the sensing cost in the detection problem and assumes that the decision-maker chooses the same fixed number of processes at every time instant. Our problem setting is different from these models. Specifically, the decision-maker can choose any number of processes at each time instant, and this choice is determined by the cost associated with the observations. We also account for the potential statistical dependence among the processes.

Our anomaly detection problem is different from the sequential parameter estimation.  Sequential parameter estimation refers to estimating the random parameter of a process~\cite{yaacoub2018optimal, grambsch1983sequential,bickel1967asymptotically}. Although this goal is similar to ours, the controlled sequential selection of processes makes our problem fundamentally different from sequential parameter estimation. In particular, only one Bernoulli process is considered in~\cite{yaacoub2018optimal, grambsch1983sequential,bickel1967asymptotically}, and  at every time instant, the decision-maker only decides whether or not to continue collecting observations. Hence, the results in these studies apply to our setting only if we consider the set of all $N$ processes as a single random process with $2^N$ states and choose the sub-optimal strategy of observing all the processes all the time. 
\subsection{Our Contributions}\label{sec:contributions}
{To the best of our knowledge, ours is the first work that formulated the anomaly detection problem as an active hypothesis testing problem and developed specific solutions for the problem.} Our specific contributions are as follows:

\begin{itemize}

\item \emph{Formulation of anomaly detection as a Markov decision process:} In \Cref{sec:mdp}, we formulate the anomaly detection problem as a Markov decision process (MDP). We define the posterior belief vector on the conditions of the processes as the state of the MDP, the  subset of processes chosen by the decision-making agent as the action, and two different types of reward functions: an average Bayesian log-likelihood ratio (LLR) based reward and an entropy-based reward. The rewards are designed such that the optimal policy of the MDP minimize the sensing cost and the delay in decision-making. Finally, the process/sensor selection is formulated as the problem of  finding an optimal policy that maximizes the long-term reward of the MDP subject to the condition that the confidence level on the estimate exceeds a specific value.
\item \emph{RL algorithms:} In \Cref{sec:actorcritic}, we develop the deep RL framework through which an optimal policy that maximizes the discounted cumulative reward is learned. We develop two deep RL algorithms, based on the Q-learning and actor-critic frameworks.

\item \emph{Active inference:} In \Cref{sec:active_infer}, we present an alternative solution strategy called active inference.  Here, we define the notion of \emph{free-energy} based on the entropy associated with the estimate of the states of the processes and the sensing cost and reformulate the anomaly detection problem as an active inference problem to minimize the free energy. The resulting algorithm is implemented using deep neural networks which are relatively less explored for active inference.

\item \emph{Empirical validation:}
Via our numerical results presented in \Cref{sec:simulations}, we investigate the performance of different frameworks and algorithms in terms of detection accuracy, delay, and sensing cost. We show that the active inference algorithm is more robust to the variations in the system parameters and adapts better to statistical dependence among the processes. Further, we observe that as the statistical dependence between the states of the
processes increases, the delay in state estimation gets diminished. This result implies that unlike the traditional Chernoff test, our algorithms are able to learn and exploit any underlying statistical dependence among the processes to reduce the number of observations.
\end{itemize}
In summary, we use the model-based posterior updates to tackle the uncertainties in the observations and the data-driven neural networks to handle the underlying statistical dependence between the processes, balancing the model-based and the data-driven approaches. 

Furthermore, in this paper, compared to the conference versions, we conduct a more comprehensive and unified analysis of deep learning-based anomaly detection and make several new contributions: 1) We design a new RL algorithm based on the deep Q-learning algorithm which we implement using the dueling architecture; 2) in addition to the LLR-based reward, we introduce an entropy-based reward (newly applied in deep RL algorithms), and we mathematically show that the two reward functions encourage the agent to achieve the desired confidence level as quickly as possible (see \Cref{prop:compare}); 3) we derive the Chernoff test for the anomaly detection problem and compare its performance with our algorithms; 4) we present a detailed numerical study that compares the different algorithms when the cost and flipping probabilities are different across the processes.

The remainder of the paper is organized as follows. We present the system model in \Cref{sec:anomaly} and describe the MDP problem in \Cref{sec:mdp}. In \Cref{sec:actorcritic,sec:active_infer}, we present our RL and active inference algorithms, respectively. We provide the simulation results in \Cref{sec:simulations} and offer our concluding remarks in \Cref{sec:con}.
%

\section{Anomaly Detection Problem}\label{sec:anomaly}
We consider a set of $N$ processes wherein each process is in one of the two conditions: normal (denoted by 0) or anomalous (denoted by 1). The condition of the $i\nth$ process is denoted by the $i\nth$ entry $\vecx_i$ of a random vector $\vecx\in\{0,1\}^N$. {The vector $\vecx$ can take $M\triangleq 2^N$ possible values denoted by $\lc \vech^{(i)}, \; i=1,2,\ldots,M\rc$. The conditions of these processes (entries of $\vecx$) can be potentially statistically dependent. This dependence is captured by the prior distribution of $\vecx$ that is denoted using $\vecpi(0)\in [0,1]^{M}$ whose $i\nth$ entry $\vecpi_i(0)$ is 
\begin{equation}\label{eq:prior}
\vecpi_i(0) = \bbP\lc\vecx=\vech^{(i)}\rc.
\end{equation} }

Our goal is to identify the anomalous processes out of the $N$ processes, which is equivalent to estimating the random vector $\vecx$. To estimate $\vecx$, the decision-making agent probes one or more processes at every time instant and obtains potentially erroneous observations of the corresponding entries of $\vecx$. Let the set of processes probed at time $t$ be $\calA(t)\in\calP(N)$ where $\calP(N)$ denotes the power set of $\{1,2,\ldots,N\}$ without the null set (i.e., $\lv\calP(N)\rv=M-1$). Also, let the observation corresponding to the $i\nth$ process at time $t$ be denoted as $\vecy_i(t)$. Depending on the condition, $\vecy_i(t)$ obeys the following probabilistic model:
\begin{equation}\label{eq:mesurement}
\vecy_{i}(t) = \begin{cases}
\vecx_{i} & {\text{ with probability } 1-p_i}\\
1-\vecx_{i}& {\text{ with probability } p_i},
\end{cases}
\end{equation}
where $p_i\in [0,0.5]$ is called the cross-over (or flipping) probability of the $i\nth$ process. We also assume that given $\vecx$, the observations are jointly (conditionally) independent:
\begin{equation}\label{eq:indepen}
\bbP\ls\lc\vecy(\tau)\rc_{\tau=1}^t\middle|\vecx\rs = \prod_{\tau=1}^{t}\prod_{k=1}^{N}\bbP\ls\vecy_{k}(\tau)\middle|\vecx\rs,
\end{equation}
for any $t>0$ where $\vecy(\tau)\in\{0,1\}^N$ and its $k\nth$ entry is $\vecy_k(\tau)$. Further, probing the $i\nth$ process incurs a cost of $c_i>0$.  In short, at every time instant $t$, the decision-maker probes the processes indexed by $\calA(k)$, obtains the corresponding observations denoted by $\vecy_{\calA(t)}(t)\in\{0,1\}^{\lv\calA(t)\rv}$, and incurs a sensing cost of $\sum_{k\in\calA(t)} c_k$.

In this setting, the three performance metrics associated with the decision making are the following:
\begin{enumerate}
\item \emph{stopping time} denoted by $T$ which is the time instant when the decision-maker ends the observation acquisition phase and yields its estimate $\hat{\vecx}$ of $\vecx$;
\item \emph{detection accuracy} given by the conditional probability  $\bbP\ls\hat{\vecx} = \vecx \middle| \lc\vecy_{\calA(t)}(t)\rc_{t=1}^{T} \rs$;
\item \emph{sensing cost} given by $\sum_{t=1}^T\sum_{k\in\calA(t)} c_k$ which represents the total cost incurred during the observation acquisition.
\end{enumerate}
The strategy of probing all the processes at all times may lead to the most accurate and fastest decision, but at the expense of a higher sensing cost. Therefore, the decision-maker sequentially chooses a subset of processes $\calA(t)\in\calP(N)$ to balance the trade-offs between the three performance metrics. Our decision-making algorithm has two components:
\begin{enumerate}
\item  \emph{sequential process selection:} a mechanism to choose $\calA(t)\in\calP(N)$ at every time $t$,
\item \emph{stopping rule:} a mechanism to determine when to stop taking observations and declare $\hat{\vecx}$.
\end{enumerate}

We next present a novel anomaly detection algorithm that we derive by casting the detection as a learning problem using an MDP framework.

\section{MDP-based Learning Problem Formulation}\label{sec:mdp}

\begin{figure}[t]
\begin{center}

\tikzstyle{cloud} = [draw=none,circle,text centered,scale = 1,text width = 0.7cm]
\tikzstyle{cloud2} = [cloud, fill=colororange!30,node distance = 1.5cm]
\tikzstyle{cloud1} = [cloud,rectangle, fill=coloryellow!30,node distance = 1.85cm,text width = 1.3cm]
\tikzstyle{cloud3} = [cloud, fill=colorgreen!30,node distance = 1.7cm]
\tikzstyle{cloud4} = [cloud, fill=colorblue!30,node distance=1.5cm]
\tikzstyle{line} = [->,line width=1pt]
\begin{tikzpicture}

  \node[cloud1] (y1) {$\vecy_{\calA(1)}(1)$};
  \node[cloud2,below of = y1] (a1) {$\calA(1)$};
  \node[cloud3,below of = a1] (p1) {$\vecpi(1)$};

  \node[cloud3,left of = p1] (p0) {$\vecpi(0)$};

  \node[cloud1,right of = y1]   (y2) {$\vecy_{\calA(2)}(2)$};
  \node[cloud2,below of = y2] (a2) {$\calA(2)$};
  \node[cloud3,below of = a2] (p2) {$\vecpi(2)$};

  \node[cloud1,right of = y2,node distance = 3.3cm]   (yK) {$\vecy_{\calA(T)}(T)$};
  \node[cloud2,below of = yK] (aK) {$\calA(T)$};
  \node[cloud3,below of = aK] (pK) {$\vecpi(T)$};

  \node at ($(y2)!.5!(yK)$)(y) {\ldots};    	
  \node at ($(a2)!.5!(aK)$) {\ldots};
  \node at ($(p2)!.5!(pK)$) (p) {\ldots};

  \node[cloud4,above of = y2] (s) {$\vecx$};

  \draw [line] (p0) -- (p1) ;
  \draw [line] (p1) -- (p2) ;
  \draw [line] (p2) -- (p) ;
   \draw [line] (p) -- (pK);

  \draw [line] (a1) -- (p1) ;
  \draw [line] (a2) -- (p2) ;
  \draw [line] (aK) -- (pK) ;

  \draw [line] (a1) -- (y1) ;
  \draw [line] (a2) -- (y2) ;
  \draw [line] (aK) -- (yK) ;

  \draw [line] (s) -- (y1) ;
  \draw [line] (s) -- (y2) ;
  \draw [line] (s) -- (yK) ;
\end{tikzpicture}

\end{center}
\caption{Graphical model with posteriors (MDP states), actions and observations}
\label{fig:pgm}

\end{figure}
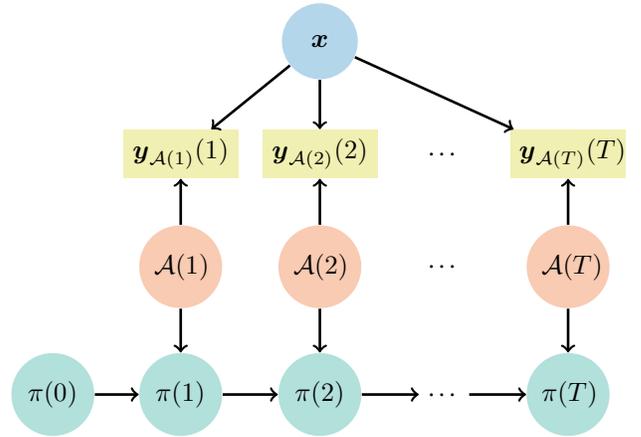

This section describes the MDP framework that models the anomaly detection problem. An MDP represents a sequential decision making problem in stochastic environments wherein the state of the environment depends on the action of a decision-maker. The goal of the decision-maker is to sequentially decide which action to choose while in a given state to maximize the reward which is the same as finding a mapping from the states to the actions. This mapping is referred to as a policy and it can be either deterministic (i.e., a one-to-one mapping) or stochastic (described by conditional probability distributions over actions given the states). 


\subsection{State and Action}
 In the context of our anomaly detection problem, we define the state of the MDP at time $t$ as the posterior belief vector $\vecpi(t)$ on the random vector
$\vecx\in\{0,1\}^N$.  The $i\nth$ entry of the posterior belief vector $\vecpi(t)\in[0,1]^{M}$ is defined as
\begin{equation}\label{eq:posterior_defn}
\vecpi_i(t) = \bbP\ls\vecx = {\vech^{(i)}} \middle| \lc\vecy_{\calA(\tau)}(\tau)\rc_{\tau =1}^{t}\rs.
\end{equation}
Further, the action at time $t$ refers to the set of processes to be observed, $\calA(t)\in\calP(N)$.

We next establish the connections between the states, actions, and  corresponding observations which can be  represented using a probabilistic graphical model depicted in \Cref{fig:pgm}. We first note that at time $t$, the data available to the agent is $\lc \vecy_{\calA(\tau)}(\tau), \tau=1,2,\ldots,{k}\rc$ using which the posterior belief vector $\vecpi(t)\in[0,1]^{M}$ can be computed in closed form.  From \eqref{eq:posterior_defn}, $\vecpi_i(t)$ is computed recursively from \eqref{eq:indepen} and \eqref{eq:prior} as 
\begin{equation}
\vecpi_i(t) = \frac{\displaystyle\vecpi_i(t-1) \prod_{k\in\calA(t)}\bbP\ls \vecy_{k}(t)\middle| \vecx={\vech^{(i)}} \rs}
{\displaystyle\sum_{j=1}^{M} \vecpi_{j}(t-1) \prod_{k\in\calA(t)} \bbP\ls \vecy_{k}(t)\middle| \vecx={\vech^{(j)}} \rs},\label{eq:posterior_update}
\end{equation}
where we obtain from \eqref{eq:mesurement} that
\begin{multline}
\bbP\ls \vecy_{k}(t)\middle| \vecx={\vech^{(i)}} \rs\\= (1-p_k)\mathbbm{1}_{\lc\vecy_{k}(t) = {\vech_k^{(i)}}\rc}+ p_k\mathbbm{1}_{\lc\vecy_{k}(t) \neq {\vech_k^{(i)}}\rc},
\end{multline} 
$\mathbbm{1}$ is the indicator function, and ${\vech_k^{(i)}}\in\{0,1\}$ is the $k\nth$ entry of ${\vech^{(i)}}$. 
As a result, given the previous posterior $\vecpi(t-1)$,  the action $\calA(t)$ and the observation $\vecy_{\calA(t)}$, we can exactly compute the updated posterior belief vector $\vecpi(t)$ using \eqref{eq:posterior_update}. 

Before we define the notion of reward, we recall from \Cref{sec:anomaly} that our goal is to achieve a balance between the detection accuracy, stopping time, and sensing cost. We can use the posteriors or the MDP states to control the detection accuracy via the stopping rule using a parameter $\upi\in(0,1)$. The parameter $\upi$ represents the threshold on the largest value among the beliefs on the different values of $\vecx$,
\begin{equation}\label{eq:stoppingrule}
\underset{i=1,2,\ldots,m}{\max} \; \vecpi_i(T)\geq \upi.
\end{equation}
Having defined the stopping rule (one of the two components of the algorithm), which controls the detection accuracy, we next focus on the trade-off between the stopping time and sensing cost. This trade-off is determined by the sequential process selection (the other algorithm component), which we derive using the notion of reward and policy.

\subsection{Reward}\label{sec:reward}
The reward $r(t)$ at time $t$ is a function of the posterior beliefs  $\vecpi(t)$ and $\vecpi(t-1)$, and the selected processes $\calA(t)$. The reward indicates the intrinsic desirability of choosing the subset of processes as  a function of the posterior belief. For a given reward function, the policy $\mu:[0,1]^{M}\to\calP(N)$ is a mapping from the posterior belief vector $\vecpi(t-1)$ to the processes to be probed $\calA(t)$.  The policy represents the sequential process selection part of our algorithm, and it is designed to maximize the long-term reward given as follows:
\begin{equation}\label{eq:long-term1}
R(T) =\sum_{t=1}^T\expect{}{r(t)}.
\end{equation}
Here, the expectation is over the uncertainty in the value of $\vecx$ and the observations $\vecy_{\calA(t)}(t)$ when $\calA(t)$ follows the policy~$\mu$. 

The reward function balances the trade-off between the stopping time and sensing cost. Here, the sensing cost can be quantified as a function of $\calA(t)$ using the term $\sum_{k\in\calA(t)}c_k$. However, we also need a term in the reward which forces the policy to build the posterior belief on the true value of $\vecx$ as quickly as possible, and thus minimize the stopping  time $T$. We represent this term using $\xi:[0,1]^{M}\to\bbR$ which is a function of the posterior beliefs. We seek $\xi$ that encourages the decision-maker to move away from the non-informative posterior $\vecpi(t) = 1/M\one$ and move towards the posterior $\vecpi(t) \in \lc\vece_i\rc_{i=1}^{M} $. Here, $\one$ denotes the all-ones vector and $\vece_i\in\{0,1\}^{M}$ denotes the $i\nth$ column of the $M\times M$ identity matrix. Two functions that achieve this goal are given by the following proposition{~\cite{cover1999elements}}.
\begin{prop}\label{prop:compare}
Let $L,H:[0,1]^{M}\to\bbR$ be two functions defined, respectively, as
\begin{align}\label{eq:L_defn}
L(\vecpi) &= \sum_{i=1}^{M}\vecpi_i\log\frac{\vecpi_{i}}{1-\vecpi_{i}}\\
H(\vecpi) &= -\sum_{i=1}^{M}\vecpi_i\log\vecpi_i.\label{eq:H_defn}
\end{align}
These functions satisfy
\begin{align*}
\underset{\substack{\vecpi\in [0,1]^{M}\\\sum_{i=1}^{M}\vecpi_i=1}}{\arg\min}\;L(\vecpi) &= \underset{\substack{\vecpi\in [0,1]^{M}\\\sum_{i=1}^{M}\vecpi_i=1}}{\arg\max}\;H(\vecpi)=\frac{1}{M}\one\\
\underset{\substack{\vecpi\in [0,1]^{M}\\\sum_{i=1}^{M}\vecpi_i=1}}{\arg\max}\;L(\vecpi) &= \underset{\substack{\vecpi\in [0,1]^{M}\\\sum_{i=1}^{M}\vecpi_i=1}}{\arg\min}\;H(\vecpi)=\lc\vece_i\rc_{i=1}^{M},
\end{align*}
where $\one$ denotes the all-ones vector and $\vece_i\in[0,1]^{M}$ denotes the $i\nth$ column of the $M\times M$ identity matrix.
\end{prop}
\begin{proof}
From the log-sum inequality, we have
\begin{equation*}
\sum_{i=1}^{M}a_i\log\lb\frac{a_i}{b_i}\rb\geq \lb\sum_{i=1}^{M}a_i\rb\log\lb\frac{\sum_{i=1}^{M}a_i}{\sum_{i=1}^{M}b_i}\rb,
\end{equation*}
for any set $\lc a_i\geq 0,b_i\geq 0\rc_{i=1}^{M}$,  and equality holds only if $a_i=\alpha b_i$, for some constant $\alpha>0$.
Thus, for any $\vecpi\in[0,1]^{M}$,
\begin{align*}
L(\vecpi)-L\lb \frac{1}{M}\one\rb 
&= \sum_{i=1}^{M}\vecpi_i\log\frac{\vecpi_{i}(M-1)}{1-\vecpi_{i}}\\
&\geq \lb\sum_{i=1}^{M}\vecpi_i\rb\log\lb\frac{\sum_{i=1}^{M}\vecpi_i}{\sum_{i=1}^{M}\frac{1-\vecpi_i}{M-1}}\rb=0.
\end{align*}
Hence, $L(\vecpi)\geq L\lb \frac{1}{M}\one\rb$ and equality holds only if $\vecpi_i=1/M$.

We next look at the maximum of $L(\vecpi)$ and we see that if $\vecpi_i\in[0,1)$ for all values of $i$, $L(\vecpi)<\infty$. Therefore, $L(\vecpi)$ attains the maximum value if and only if at least one entry of $\vecpi$ is 1. Hence, the desired maxima is achieved at $\lc\vece_i\rc_{i=1}^{M}$.

We can compute the maxima and minima of $H(\vecpi)$ using similar arguments and thus, the proof is complete.
\end{proof}

The above proposition implies that the functions $L$ and $H$ are two good choices for $\xi$. We note that in \eqref{eq:L_defn}, the term $\log\frac{\vecpi_i}{1-\vecpi_i}$ is the likelihood ratio (LLR) of the two hypotheses namely, $\vecx={\vech^{(i)}}$ and $\vecx\neq{\vech^{(i)}}$. Consequently, $L(\vecpi)$ is the Bayesian LLR obtained
by applying the logit function on the posterior belief. Further, maximizing the Bayesian LLR increases the posterior belief on the true value of $\vecx$. Also, $H$ is the entropy of the distribution $\vecpi$, and thus, minimizing $H$ reduces the uncertainty in estimation and builds the posterior belief on the true value of $\vecx$.

Having defined the function\footnote{The algorithmic development is independent of the choice of the reward function (LLR and entropy-based). Therefore, in the remainder of the paper,  we use $\xi(\cdot)$ in the reward of the MDP which can either be $L(\cdot)$ defined in \eqref{eq:L_defn} or $-H(\cdot)$ defined in \eqref{eq:H_defn}. } $\xi$, we formulate the instantaneous reward of the MDP as a weighted sum of $\xi$ and the sensing cost:
\begin{equation}\label{eq:reward}
r(t)= \xi(\vecpi(t))-\xi(\vecpi(t-1))+\lambda \sum_{k\in\calA(t)} c_k,
\end{equation}
where $\lambda>0$ is the weighing parameter that dictates the balance between the stopping time and the total sensing cost. Thus, from \eqref{eq:long-term1} and \eqref{eq:reward}, the long-term expected reward of the MDP up to time $t$ is given by
\begin{equation*}
R(t) = \expect{}{\xi(\vecpi(t)) - \xi(\vecpi(0))-\lambda\sum_{\tau=1}^t\sum_{k\in\calA(\tau)} c_k},
\end{equation*}
{where the expectation is over the distribution of $\vecx$ and the observations $\vecy_{\calA(t)}(t)$ given $\calA(t)$.}
The MDP objective is to find a policy or sequence of actions $\lc\calA(t)\in\calP(N)\rc_{t=1}^T$ that maximizes the long-term average sum of the rewards. Hence, a policy that maximizes the long-term reward improves the accuracy of the estimate (quantified by $\xi(\vecpi(t))$) as soon as possible while minimizing the overall sensing cost (the last term in $R(t)$). Further, the agent continues to take observations until it declares an estimate with the desired level of confidence given by $\upi$ (i.e., $t=T$). Therefore, if $\lambda$ is small, the reward ensures that the agent chooses actions with a significant change $\xi(\vecpi(t)) - \xi(\vecpi(0))$, leading to a shorter stopping time. On the other hand, with a large $\lambda$, the agent tries to minimize the sensing cost by probing a few processes at every time instant which increases the stopping time. Therefore, $\lambda$ controls the stopping time and total sensing cost. 

Further, the Bayesian LLR $L$ is unbounded unlike the entropy satisfying $H(\vecpi)\leq \log M$ for any $\vecpi\in[0,1]^{M}$. Also, 
\begin{equation}\label{eq:compare_CH}
L(\vecpi)=-H(\vecpi)-\sum_{i=1}^{M}\vecpi_i\log(1-\vecpi_i)\geq -H(\vecpi).
\end{equation}
Therefore, for the same value of $\lambda$, the Bayesian LLR reward function gives a higher weight to the accuracy than the cost. As a result, the sensitivity of the trade-off between the accuracy and sensing cost differs for the two reward functions. We discuss this point in detail in \Cref{sec:simulations}.

This completes our discussion on the reward function. Using this formulation, we next present the deep learning algorithms to obtain policies that maximize the long-term reward of the MDP. We use two approaches: the deep RL-based approach presented in \Cref{sec:actorcritic} and deep active inference-based approach presented in \Cref{sec:active_infer}.
\section{Anomaly Detection Using Deep RL Algorithms}\label{sec:actorcritic}
Our RL algorithms are designed to maximize the expected discounted return $\bar{R}(t)$ defined as
\begin{equation}\label{eq:Rt_defn}
\bar{R}(t) = \lim_{T\to\infty}\sum_{j=0}^T \gamma^{j} r(t+j),
\end{equation}
where $0<\gamma <1$ (which is generally close to 1) is the discount factor. This parameter $\gamma$ weighs the rewards in the distant future relative to those in the immediate future, i.e., a reward received $j$ time steps in the future is worth only $\gamma^{j-1}$ times what it would be worth if it were received immediately. Hence, this approach encourages the agent to minimize the stopping time.

To maximize $\bar{R}(t)$, the RL algorithms make process selection based on the value functions of the posterior-action pair and the posterior. For a given policy $\mu$, these value functions are
\begin{align}
Q_{\mu}(\vecpi,\calA) &= \expect{}{\bar{R}_k\middle| \vecpi(t-1)=\vecpi,\calA(t)=\calA}\label{eq:Q_defn}\\
V_{\mu}(\vecpi) &= \expect{\calA\sim\mu(\vecpi)}{\bar{R}_k\middle| \vecpi(t-1)=\vecpi},\label{eq:value_defn}
\end{align}
where the expectations are evaluated given that the agent follows the policy $\mu$ for all future actions.
Intuitively, the action-value function (referred to as the $Q$-function), $Q_{\mu}(\vecpi,\calA)$, in \eqref{eq:Q_defn} indicates the long term desirability of choosing a particular action when the posterior belief vector is $\vecpi$. Also, the state-value function (referred to as the value function), $V_{\mu}(\vecpi)$, in \eqref{eq:value_defn} specifies the expected reward when starting with posterior belief vector $\vecpi$ and following the policy $\mu$ thereafter. An RL agent makes the action choices by evaluating the optimal value estimates, $Q$-function or the value function, or both. If we have the optimal values of the functions, then the actions that appear best after a one-step search are the optimal actions~\cite{sutton2018reinforcement}. In the following, we present two different RL approaches, the Q-learning and actor-critic algorithms, and describe how they estimate these functions to arrive at the optimal policy.

\subsection{Dueling Deep Q-learning}

The Q-learning approach is a popular RL algorithm where the agent estimates the $Q$-function and chooses the action $\calA(t)$ that maximizes the $Q$-function given the posterior belief vector $\vecpi(t-1)$~\cite{watkins1989learning}. Further, in the deep Q-learning framework, the unknown $Q$-function is modeled using a neural network~\cite{mnih2015human}, and the dueling deep Q-learning framework refers to the implementation of this neural network using a model called the dueling architecture~\cite{wang2016dueling}.

 This architecture consists of a single $Q$-network and relies on a quantity called the advantage function: $
A_{\mu}(\vecpi,\calA) = Q_{\mu}(\vecpi,\calA)-V_{\mu}(\vecpi)$, which is a measure of how much $Q_{\mu}(\vecpi,\calA)$ deviates from the expected value over all the actions, $V_{\mu}(\vecpi)$, and therefore,  specifies the relative preference of each action for a given posterior belief vector. The dueling architecture estimates both $V_{\mu}(\vecpi)$ and $A_{\mu}(\vecpi,\calA)$ separately and combines them to obtain $Q_{\mu}(\vecpi,\calA)$. The input to the $Q$-network is the posterior belief vector $\vecpi\in\bbR^{M}$ and the output is an $(M-1)$-length vector whose $i\nth$ entry corresponds to $Q_{\mu}(\vecpi,\cdot)$ of the $i\nth$ possible action. The network parameter $\theta_{\mathrm{DQN}}$  is obtained by optimizing the loss function~\cite{wang2016dueling}:
\begin{multline}\label{eq:Qloss}
L_{\mathrm{DQN}}(\theta_{\mathrm{DQN}}) \!=\! \mathbb{E}_{\vecpi,\calA,\vecpi'}\Bigg\{\!r(t) + \gamma\underset{\calA'\in\calP(N)}{\max}Q(\vecpi',\calA';\theta_{\mathrm{DQN}}^-) \\ - Q(\vecpi,\calA;\theta_{\mathrm{DQN}})\Bigg\},
\end{multline}
where $\theta_{\mathrm{DQN}}^-$ is the current network parameter estimate obtained in the previous time. We update the $Q$-function using bootstrapping by basing its update in part on an current estimate $Q(\vecpi',\calA';\theta_{\mathrm{DQN}}^-)$. Also, to ensure that the learned value of the $Q$-function converges to the optimal $Q$-function, we use the greedy policy to choose the successor action~$\calA'$.

Using the learned $Q$-function, we next describe how to obtain the optimal policy. We derive the policy using a combination of the decaying-epsilon greedy algorithm{~\cite{wang2016dueling,ostovar2018adaptive}} and the Gibbs softmax method{~\cite{kong2020rankmax}}. At every time step, the agent takes action using the Gibbs softmax method with a probability of $1-\epsilon$ and a random action with a probability of $\epsilon$. Here, $\epsilon$ is a parameter that decays with time. Also, the Gibbs softmax method refers to choosing the action $\calA(t) \sim\sigma\lb Q(\vecpi(t),\cdot;\theta_{\mathrm{DQN}})\rb\in[0,1]^{M-1}$ where $\sigma(\cdot)$ is the softmax function. The approach ensures that the entire action space is explored while exploiting the best action with high probability.

The algorithm chooses actions in the above fashion until the posterior belief distribution satisfies the stopping rule in \eqref{eq:stoppingrule}. We present the pseudo-code for our dueling deep Q-learning-based detection algorithm in \Cref{alg:deep Q-learning}.

\begin{algorithm}[hpt]
\caption{Dueling Q-learning algorithm for anomaly detection}
\label{alg:deep Q-learning}
\begin{algorithmic}[1]
\REQUIRE Prior distribution $\vecpi(0)$, discount rate $\gamma\in(0,1)$, and confidence level $\upi\in(0,1]$

\ENSURE $Q$-network parameter $\theta_{\mathrm{DQN}}$ arbitrarily

\REPEAT
\STATE Time index $t =1$
\WHILE {$\underset{i}{\max} \; \vecpi_i(t-1) < \upi$}
\STATE Choose action $\calA(t)$ using the policy derived from $Q(\vecpi(t),\cdot;\theta_{\mathrm{DQN}})$
\STATE Generate observations $\vecy_{\calA(t)}(t)$
\STATE Compute $\vecpi(t)$ using \eqref{eq:posterior_update} and $r(t)$ using  \eqref{eq:reward}
\STATE Update $\theta_{\mathrm{DQN}}$ by minimizing $L_{\mathrm{DQN}}(\theta_{\mathrm{DQN}})$ in \eqref{eq:Qloss}
\STATE Increase time index $t=t+1$
\ENDWHILE
\STATE Declare  $\hat{\vecx}={\vech_k^{(i^*)}}$ where $i^*=\underset{i}{\arg\max} \; \vecpi_i(t-1)$
\UNTIL {}
\end{algorithmic}
\end{algorithm}

\subsection{Deep Actor-critic}
The deep actor-critic algorithm is another RL algorithm that directly learns the policy. This principle differs from that of the dueling deep Q-learning algorithm, which learns the $Q$-function and derives a policy based on the learned $Q$-function.

The actor-critic architecture consists of two separate neural networks, actor and critic, with no shared features. The actor learns the policy, which chooses the action based on the posterior probabilities. Thus, its input is $\vecpi\in\bbR^{M}$ and the output is $\mu_{\mathrm{AC}}(\vecpi,\cdot;\theta_{\mathrm{actor}})\in[0,1]^{M-1}$ where $\theta_{\mathrm{actor}}$ represents the neural network parameters. The policy returned by the actor network is a stochastic policy which chooses an action according to $\calA\sim\mu_{\mathrm{AC}}(\vecpi,\cdot;\theta_{\mathrm{actor}})$. The critic refers to the learned value function, which is an estimate of how good the policy learned by the actor is and hence, essentially provides an evaluation of that policy. The evaluation of the action $\calA(t)$ taken corresponding to the posterior $\vecpi(t-1)$ takes the form of the temporal-difference (TD) error as given by 
\begin{equation*}
\delta(t) = r(t)+\gamma V_{\mu}(\vecpi(t))-V_{\mu}(\vecpi(t-1)),
\end{equation*}
for a given policy $\mu(\cdot)$ with
\begin{equation*}
V_{\mu}(\vecpi) = \expect{\vecpi',\calA\sim\mu(\vecpi)}{r(t)+ \gamma V_{\mu}(\vecpi')\middle| \vecpi(t-1)=\vecpi}.
\end{equation*}
 If the TD error is positive, the probability of choosing $\calA(t)$ in the future is increased, and vice versa. Therefore, the input to the critic network is the posterior belief vectors $\vecpi\in[0,1]^{M}$ and the output is the corresponding value function $V(\vecpi;\theta_{\mathrm{critic}})\in\bbR$ where $\theta_{\mathrm{critic}}$ represents the neural network parameters. 

We next describe how to learn the two sets of network parameters: $\theta_{\mathrm{actor}}$ of the actor and $\theta_{\mathrm{critic}}$ of the critic. Since the goal of the critic network is to fit a model to estimate the optimal value function, its parameter update is equivalent to minimizing the model mismatch between the reward obtained at the current time step and the learned value function. Thus, the critic network updates its parameter $\theta_{\mathrm{critic}}$ by minimizing the square of the TD error given by
\begin{equation}\label{eq:delta}
\delta(t) = r(t)+\gamma V(\vecpi(t);\theta_{\mathrm{critic}})-V(\vecpi(t-1);\theta_{\mathrm{critic}}^-),
\end{equation}
where $\theta_{\mathrm{critic}}^-$ is the current critic network parameter estimate obtained in the previous time instant.
On the other hand, the goal of the actor network is to find a policy that maximizes the value function. Thus, its parameter update is via maximization of the value function. The actor updates its parameter~\cite{sutton2018reinforcement} as
\begin{equation}\label{eq:policy_gradient}
\theta_{\mathrm{actor}}=\! \theta_{\mathrm{actor}}^- + \delta(t)\nabla_{\theta_{\mathrm{actor}}} \ls \log \mu_{\mathrm{AC}}(\vecpi(t\!-\!1),\calA(t);\theta_{\mathrm{actor}})\rs,
\end{equation}
where $\theta_{\mathrm{actor}}^-$ is the current actor network parameter estimate obtained in the previous time instant and $\delta(t)$ given by \eqref{eq:delta} is obtained from the critic network.

The learned policy is straightforward in the case of the actor-critic framework, as the actor network directly learns the policy. Hence, at every time step, the agent chooses an action based on the output of the actor network and receives the reward for updating both actor and critic networks. The agent stops collecting observations and returns an estimate of $\vecx$ when the confidence level exceeds the desired level, i.e., when \eqref{eq:stoppingrule} holds. The pseudo-code of our algorithm is given in \Cref{alg:ActorCritic}.

\begin{algorithm}[hpt]
\caption{Actor-critic RL for anomaly detection}
\label{alg:ActorCritic}
\begin{algorithmic}[1]
\REQUIRE
Prior distribution $\vecpi(0)$, discount rate $\gamma\in(0,1)$, and confidence level $\upi\in(0,1]$

\ENSURE Actor and critic neural network parameters $\theta_{\mathrm{actor}}$ and $\theta_{\mathrm{critic}}$ arbitrarily

\REPEAT
\STATE Time index $t =1$
\WHILE {$\underset{i}{\max} \; \vecpi_i(t-1) < \vecpi_{\mathrm{upper}}$}
\STATE Choose action $\calA(t)$ using the policy derived from $\mu_{\mathrm{AC}}(\vecpi(t-1),\cdot;\theta_{\mathrm{actor}})$
\STATE Generate observations $\vecy_{\calA(t)}(t)$
\STATE Compute $\vecpi(t)$ using \eqref{eq:posterior_update} and $r(t)$ using  \eqref{eq:reward}
\STATE Update $\theta_{\mathrm{critic}}$ by minimizing the squared temporal error $\delta^2(t)$ in \eqref{eq:delta}
\STATE Update $\theta_{\mathrm{actor}}$ using \eqref{eq:policy_gradient}
\STATE Increase time index $t=t+1$
\ENDWHILE
\STATE Declare $\hat{\vecx}={\vech_k^{(i^*)}}$ where $i^*=\underset{i}{\arg\max} \; \vecpi_i(t-1)$
\UNTIL{}
\end{algorithmic}
\end{algorithm}

\section{Anomaly Detection Using Deep Active Inference}\label{sec:active_infer}
The active inference framework is an alternate approach to solving the MDP  problem described in \Cref{sec:mdp}. It is inspired by a normative theory of brain function based on its perception of the MDP, i.e., the active inference agent maintains a generative model that represents its perception~\cite{friston2017active,friston2017active_b,friston2015active}. This generative model $\phi(\cdot)$ comprises a joint probability distribution on the posterior, the actions, and the corresponding observations: $\lc\vecpi(t-1),\calA(t),\vecy_{\calA(t)}(t),t>0\rc$. The model assigns higher probabilities to the posteriors and actions favorable to the agent. Given a generative model, the agent inverts the model to find the conditional distribution of the action $\calA(t)$ corresponding to the posterior $\vecpi(t-1)$. However, since directly computing the marginals is difficult, we use the method of approximate Bayesian inference. To this end, it defines a variational distribution $\mu_{\mathrm{AI}}(\vecpi,\calA)$  that is controlled by the agent. The distribution $\mu_{\mathrm{AI}}(\cdot)$ is optimized by minimizing the Kullback-Leibler (KL) divergence between the distributions $\mu_{\mathrm{AI}}(\cdot)$ and $\phi(\cdot)$. Therefore, a stochastic policy that chooses actions according to the distribution $\mu_{\mathrm{AI}}(\cdot)$ maximizes the obtained reward. The KL divergence between the variational distribution and the generative model is called the variational free energy. In other words, the goal of the active inference agent is to find the stochastic policy $\mu_{\mathrm{AI}}(\cdot)$ which minimizes its expected free energy (EFE). 

From \eqref{eq:posterior_update}, we know that the posterior belief vector $\vecpi(t)$ can be exactly inferred using the knowledge of the action $\calA(t)$, observation  $\vecy_{\calA(t)}$ and posterior belief vector $\vecpi(t-1)$. This relationship, along with the Markov property, enables us to completely define the generative model using the distribution $\phi(\calA(t),\vecy_{\calA(t)}(t)|\vecpi(t-1))$. This distribution is
\begin{multline*}
\phi(\vecy_{\calA(t)}(t),\calA(t)|\vecpi(t-1)) \\ = \phi(\vecy_{\calA(t)}|\calA(t),\vecpi(t-1))\phi(\calA(t)|\vecpi(t-1)).
\end{multline*}
The generative model is biased towards high rewards, encoded into the generative model as the prior probability of the belief,
\begin{equation}\label{eq:observation_prob}
\phi(\vecy_{\calA(t)}(t)|\calA(t),\vecpi(t-1)) = \sigma\lb r(t)\rb,
\end{equation}
where we recall that $\sigma(\cdot)$ is the softmax function and $r(t)$ denotes the instantaneous reward of the MDP at time $t$.

We next complete the construction of the generative model by specifying the distribution $\phi(\calA(t)|\vecpi(t))$. Since the agent tries to minimize the total free energy of the expected trajectories into the future, it is encoded into the generative model as
\begin{equation}\label{eq:action_distribution}
\phi(\calA(t)|\vecpi(t)) = \sigma\lb-G(\calA(t),\vecpi(t-1))\rb,
\end{equation}
where $G(\cdot)$ is the total free energy of the expected trajectories into the future, and the variational free energy is the KL divergence between the variational distribution $\mu_{\mathrm{AI}}(\cdot)$ and the generative model $\phi(\cdot)$:
\begin{multline}\label{eq:F_defn}
F(t) = \sum_{\calA(t)\in\calP(N)}\mu_{\mathrm{AI}}(\vecpi(t-1),\calA(t))\\
\times\log\frac{\mu_{\mathrm{AI}}(\vecpi(t-1),\calA(t))}{\phi(\calA(t),\vecy_{\calA(t)(t)}|\vecpi(t-1))}.
\end{multline}
Thus, the agent constructs the generative model using the EFE, obtains the optimum policy by minimizing the EFE of all the paths into the future, and chooses an action that minimizes the EFE.  In other words, determining the optimal policy reduces to computing and optimizing the EFE. 

Next, we present the neural network architecture and learning of the neural network parameters. The deep active inference algorithm consists of two neural networks: the policy and EFE.  The policy network directly learns the process selection, and therefore, it takes the posterior belief vector $\vecpi(t-1)$ as the input. Its output is the stochastic selection policy $\mu_{\mathrm{AI}}(\vecpi(t-1),\cdot;\theta_{\mathrm{policy}}) \in[0,1]^{M-1}$  which is a probability distribution on $\calP(N)$. Here, $\theta_{\mathrm{policy}}$ denotes the neural network parameters. The EFE network represents EFE's learned value, which estimates how close the learned policy is to the generative model. Thus, the input of the EFE network is the posterior $\vecpi\in[0,1]^{M}$, and the output is the EFE value $G(\vecpi,\cdot;\theta_{\mathrm{EFE}})\in\bbR^{M-1} $,  representing the EFE values of each action $\calA\in\calP(N)$ and the posterior $\vecpi$. Here, $\theta_{\mathrm{EFE}}$ denotes the parameters of the EFE network.

The EFE can be approximated as follows~\cite{millidge2020deep}:
\begin{equation*}
G(\calA(t),\vecpi(t-1)) \approx   \expect{}{-r(t)  +  G(\calA(t+1),\vecpi(t))},\label{eq:G_expand}
\end{equation*}
where we use \eqref{eq:observation_prob}.  Therefore, we learn the parameters of the EFE network by optimizing the model mismatch between the learned EFE value and the reward obtained:
\begin{multline}\label{eq:EFE_loss}
L_{\mathrm{EFE}}(\theta_{\mathrm{EFE}}) = \mathbb{E}\Big\{\big( G\lb\calA(t),\vecpi(t-1);\theta_{\mathrm{EFE}}\rb + r(t) \\-  G\lb\calA(t+1),\vecpi(t);\theta_{\mathrm{EFE}}^-\rb\big)^2\Big\},
\end{multline}
where $\theta_{\mathrm{EFE}}^-$ denotes the current estimate of the network parameter obtained in the previous time step and the expectation is over the action distribution $\calA(t+1)\sim\mu_{\mathrm{AI}}(\vecpi(t),\cdot;\theta_{\mathrm{policy}})$. To update the policy network, we minimize the variational free energy defined in \eqref{eq:F_defn}:
\begin{multline}
F(\theta_{\mathrm{policy}}) = \sum_{\calA\in\calP(N)}\mu_{\mathrm{AI}}(\vecpi(t-1),\calA;\theta_{\mathrm{policy}}))\\
\times \log\frac{\mu_{\mathrm{AI}}(\vecpi(t-1),\calA;\theta_{\mathrm{policy}})}{\sigma(r(t))\sigma(-G(\vecpi(t-1),\calA;\theta_{\mathrm{EFE}}))}.\label{eq:F_update}
\end{multline}
Since $r(t)$ is independent of $\theta_{\mathrm{policy}}$, the loss function is
\begin{multline}\label{eq:policy_loss}
L_{\mathrm{AI}}(\theta_{\mathrm{policy}}) = -H(\mu_{\mathrm{AI}}(\vecpi(t-1),\cdot;\theta_{\mathrm{policy}})) \\- \sum_{\calA\in\calP(N)}\mu_{\mathrm{AI}}(\vecpi(t-1),\calA)\log \sigma(G(\vecpi(t-1),\calA;\theta_{\mathrm{EFE}})),
\end{multline}
where $H(\cdot)$ is given by \eqref{eq:H_defn}.

As in the case of the actor-critic algorithm, the policy to be followed by the agent is directly obtained from the (policy) neural network output. The agent follows this policy to choose an action at every time instant, collects the corresponding reward, and updates the two neural networks using the obtained reward. The algorithm is summarized in \Cref{alg:ActiveInference}.

\begin{algorithm}[ht]
\caption{Active inference algorithm for anomaly detection}
\label{alg:ActiveInference}
\begin{algorithmic}[1]
\REQUIRE Prior distribution $\vecpi(0)$ and confidence level $\upi\in(0,1]$

\ENSURE Policy and EFE network parameters $\theta_{\mathrm{policy}}$ and $\theta_{\mathrm{EFE}}$ arbitrarily

\REPEAT
\STATE Time index $t =1$
\WHILE {$\underset{i}{\max} \; \vecpi_i(t-1) < \vecpi_{\mathrm{upper}}$}
\STATE Choose action $\calA(t)$ using the policy derived from $\mu_{\mathrm{AI}}(\vecpi(t-1),\cdot;\theta_{\mathrm{policy}})$
\STATE Generate observations $\vecy_{\calA(t)}(t)$
\STATE Compute $\vecpi(t)$ using \eqref{eq:posterior_update} and $r(t)$ using  \eqref{eq:reward}
\STATE Update $\theta_{\mathrm{EFE}}$ by minimizing $L_{\mathrm{EFE}}(\theta_{\mathrm{EFE}}) $  in \eqref{eq:EFE_loss}
\STATE Update $\theta_{\mathrm{policy}}$ by minimizing the variational free energy $L_{\mathrm{AI}}(\theta_{\mathrm{policy}})$ in \eqref{eq:policy_loss}
\STATE Increase time index $t=t+1$
\ENDWHILE
\STATE Declare $\hat{\vecx}={\vech_k^{(i^*)}}$ where $i^*=\underset{i}{\arg\max} \; \vecpi_i(t-1)$
\UNTIL{}

\end{algorithmic}
\end{algorithm}

\subsection{Comparison With RL Methods}
The active inference approach has many similarities to RL-based algorithms, such as learning probabilistic models, exploring and exploiting various actions, and efficient planning. In particular, the active inference algorithm closely resembles the policy gradient methods (for example, the actor-critic algorithm) since both approaches try to learn the policy directly. We recall that the actor-critic and the active inference methods have two separate neural networks. The actor network of the actor-critic algorithm and the policy network of the active inference algorithm learn the policy to be followed. In contrast, the other network estimates a function (TD error or EFE) used to evaluate and optimize the policy. However, the two algorithms are derived based on different principles, and the main differences between the RL framework and the active inference framework are as follows:
\subsubsection{Model-free and model-based} The traditional RL uses a model-free approach where the algorithm aims at reward maximization based on the $ Q$ function or the value function, or both. The algorithm does not try to explicitly learn the probabilistic model that governs the state transition or the generation of the observations of the MDP. However, the active inference model relies on a hierarchical generative model, which is based on variational free energy. It explicitly learns the model consisting of states, actions, and observations of the MDP. To be specific, in the most general setting, the deep active inference algorithm consists of four neural networks: policy network;
EFE network; observation network to learn the distribution of the observations given the state and action; and state transition network to learn the distribution of the next state given the previous state, action, and observation~\cite{millidge2020deep}. However, in our case, \eqref{eq:observation_prob} and \eqref{eq:posterior_update} define the distributions learned by the observation network and the state transition network. Hence, the active inference algorithm comprises only two neural networks. In other words, the active inference algorithm naturally allows the algorithm to incorporate any knowledge of the environment's statistics into the model.
\subsubsection{Policy optimization} The actor-critic algorithm uses the value functions directly to learn the policy. In contrast, the active inference algorithm relies on the generative probability distribution derived from softmax over the variational free energy as defined in \eqref{eq:action_distribution}. Therefore, the actor-critic algorithm maximizes the expected reward function in the future, whereas the active inference algorithm reduces the surprise in the future by learning the probabilistic model. Moreover, the objective function of the actor-critic algorithm depends only on the samples generated using the actions that the agent took within the episode as opposed to the active inference algorithm, which averages the objective function over all possible actions in the next step (see the summation in \eqref{eq:policy_loss}). Since the active inference algorithm computes the expected value, it may lead to reduced variance and better performance.

\subsection{Comparison With Chernoff Test}
The Chernoff test, a standard algorithm for active hypothesis testing~\cite{chernoff1959sequential}, sequentially chooses actions that build the posterior belief on the true value of $\vecx$ as quickly as possible. However, it does not take the sensing cost into account. It follows the stochastic policy  $\mu_{\mathrm{Chernoff}}$,
\begin{equation}\label{eq:chernoff}
\mu_{\mathrm{Chernoff}}(\vecpi(t-1),\cdot) 
=\!\underset{\substack{\vecq\in[0,1]^{M-1}\\\sum_{i}\vecq_i=1}}{\arg \max}
\underset{\substack{\hat{\vecx} \in\{0,1\}^{M}\\ \hat{\vecx}\neq  \bar{\vecx}(t-1)}} {\min}\;  \vecq\tran \vecd(\bar{\vecx}(t-1),\hat{\vecx}),
\end{equation}
where we define the $i\nth$ entry of $\vecd(\cdot)\in\bbR^{M-1}$ as
\begin{align*}
\vecd_i(\bar{\vecx}(t),\hat{\vecx}) &\triangleq\KL {p \lb y_{\calA_i}(t) |\vecx=\bar{\vecx}(t)\rb}{ p \lb y_{\calA_i}(t) |\vecx=\hat{\vecx}\rb}\\
\bar{\vecx}(t) &= {\vech^{(\tilde{i}^*)};\; \;\; \tilde{i}^*=\underset{i}{\arg\max} \; \vecpi_i(t)}.
\end{align*}
Here, $\mathrm{KL}$ denotes the KL divergence between the distributions, and $\calA_i$ denotes that $i\nth$ element of set $\calP(N)$. However, for any $\hat{\vecx}$, the KL divergence term is maximized when all the processes are selected. Thus, we have $\vecd_i(\bar{\vecx}(t),\hat{\vecx})\leq \vecd_1(\bar{\vecx}(t),\hat{\vecx})$ 
with $\calA_1=\{1,2,\ldots,N\}$ denoting the action of selecting all processes. Therefore, we arrive at
\begin{multline*}
\underset{\substack{\vecq\in[0,1]^{M-1}\\\sum_{i}\vecq_i=1}}{\max}
\underset{\substack{\hat{\vecx} \in\{0,1\}^{M}\\ \hat{\vecx}\neq  \bar{\vecx}(t-1)}} {\min}\;  \vecq\tran \vecd(\bar{\vecx}(t-1),\hat{\vecx}) \\
\leq 
\underset{\substack{\hat{\vecx} \in\{0,1\}^{M}\\ \hat{\vecx}\neq  \bar{\vecx}(t-1)}} {\min}\;   \vecd_1(\bar{\vecx}(t),\hat{\vecx}),
\end{multline*}
and equality holds when $\vecq=\begin{bmatrix}
1 & 0 \ldots 0
\end{bmatrix}\in[0,1]^{M-1}$. Therefore, the Chernoff test always chooses the action $\calA_1$ with probability one. This policy is expected as the Chernoff test does not optimize the sensing cost, and thus, it achieves a small stopping time while incurring a high sensing cost. On the contrary, our formulation balances the trade-off between the stopping time and sensing cost via $\lambda$ and exploits the statistical dependence in $\vecx$ modeled using $\vecpi(0)$. We corroborate this point using results in \Cref{sec:simulations} (e.g., see \Cref{fig:Chernoff}).

\section{Numerical Results}\label{sec:simulations}

\begin{figure}
\begin{center}

\begin{subfigure}[b]{\textwidth/8*3}
		{\includegraphics[width=\textwidth]{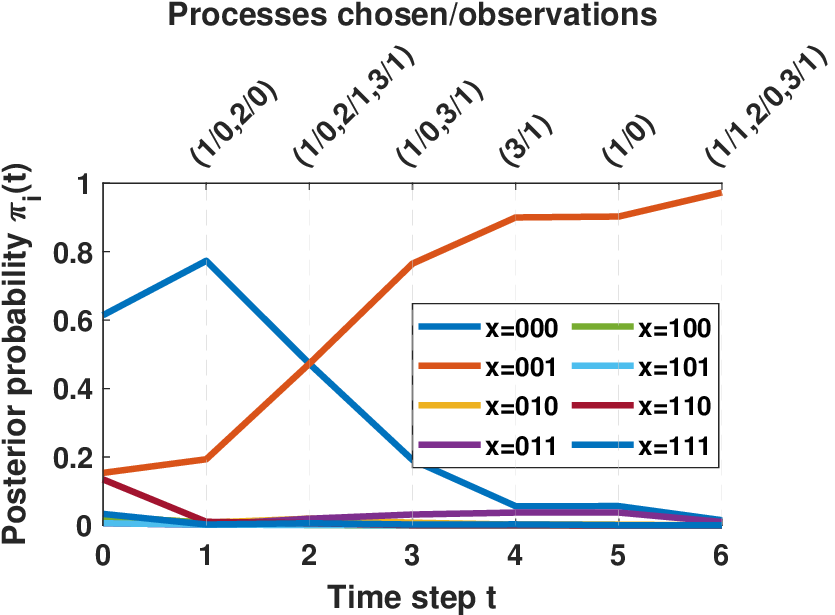}}
		\caption{{Actor-critic's policy when the process state is $[0\;0\;1]$}}
		\label{fig:demo_AC}
			\end{subfigure}
			
			\vspace{0.3cm}
			
			\begin{subfigure}[b]{\textwidth/8*3}
		{\includegraphics[width=\textwidth]{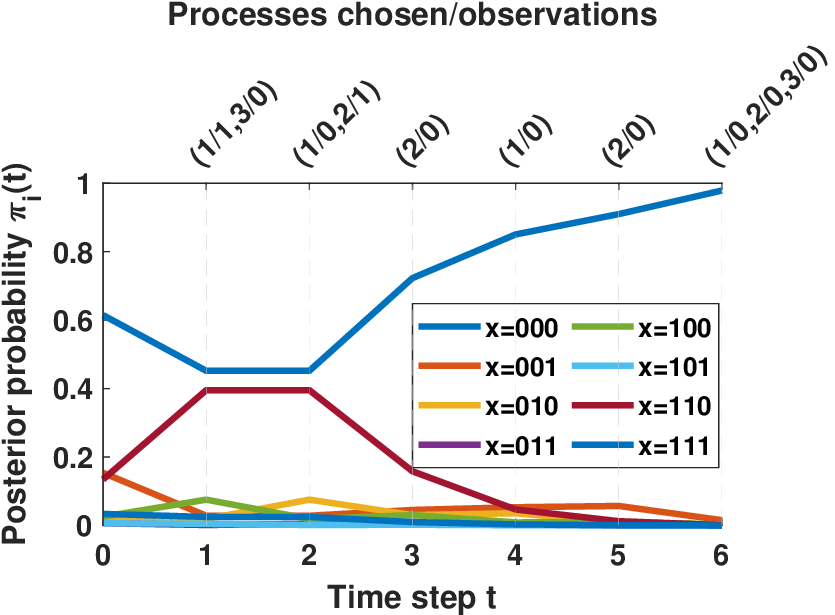}}
		
		\caption{{Active inference's policy when the process state is $[0\;0\;0]$}
		\label{fig:demo_AI}}
			\end{subfigure}
			\caption{{A single realization of the variation of the belief vector $\vecpi(t)$, sensor selection $\calA(t)$, and the corresponding observations $\vecy_{\calA(t)}(t)$ over time $t$. We choose $\upi=0.94$, $\rho=0.8$, and $\lambda=0.2$. The curves represent the evolution of the posterior probabilities of different hypotheses. The selected processes (sensors) and the corresponding observations at different times are depicted at the top of the figure.}}
			\label{fig:demo}
\end{center}
\end{figure}

\begin{figure*}[hptb]
\begin{center}

\hs{-0.3}
	\begin{subfigure}[b]{5.1cm}
		{\includegraphics[width=5.6cm]{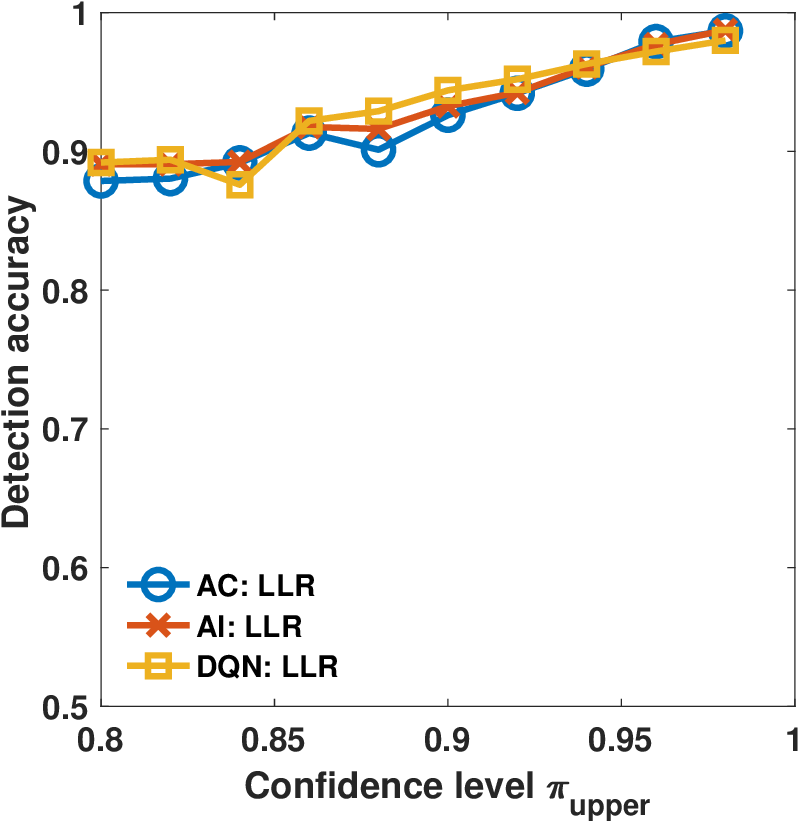}}
	 \end{subfigure}
	 \hs{0.35}
     \begin{subfigure}[b]{5.1cm}
		{\includegraphics[width=5.45cm]{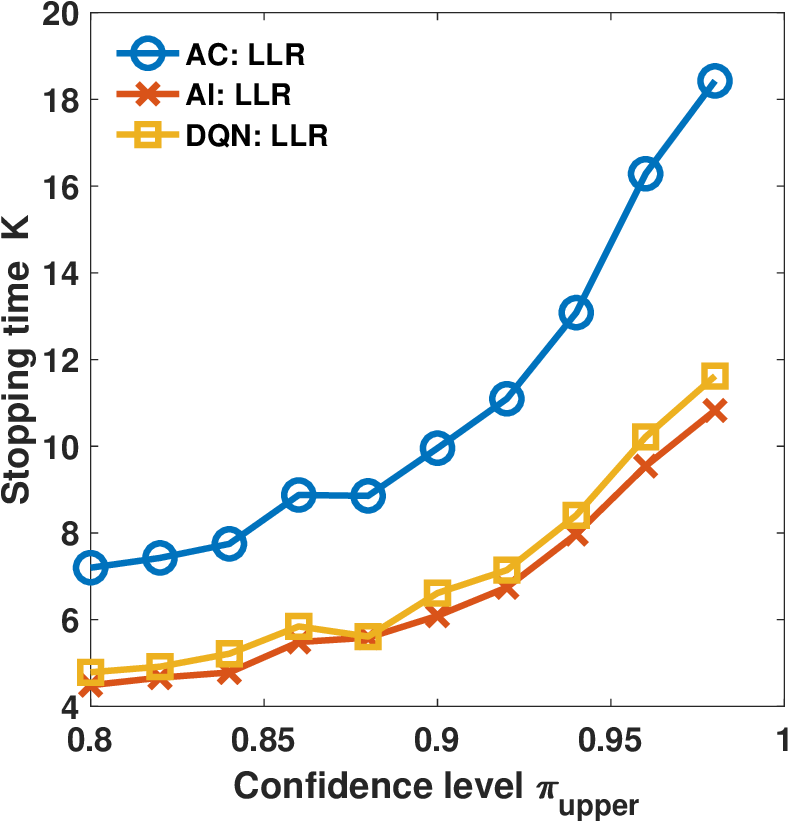}}	
      \end{subfigure}
      \hs{0.15}
	\begin{subfigure}[b]{5.1cm}
		{\includegraphics[width=5.45cm]{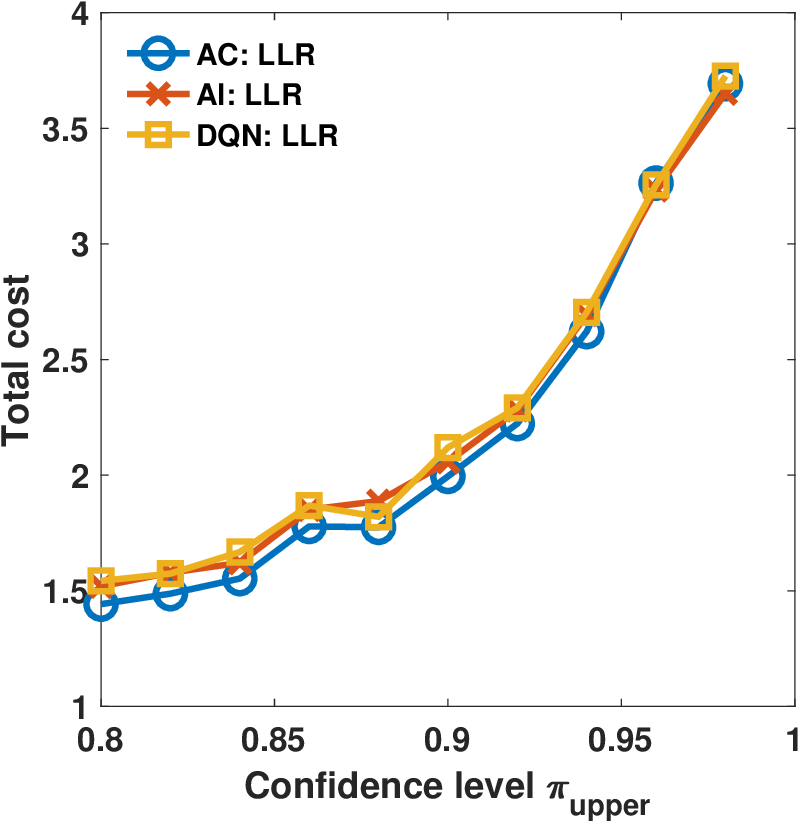}}
	\end{subfigure}
	
	\vspace{0.5cm}
	
	\hs{-0.3}
		\begin{subfigure}[b]{5.1cm}
		   {\includegraphics[width=5.7cm]{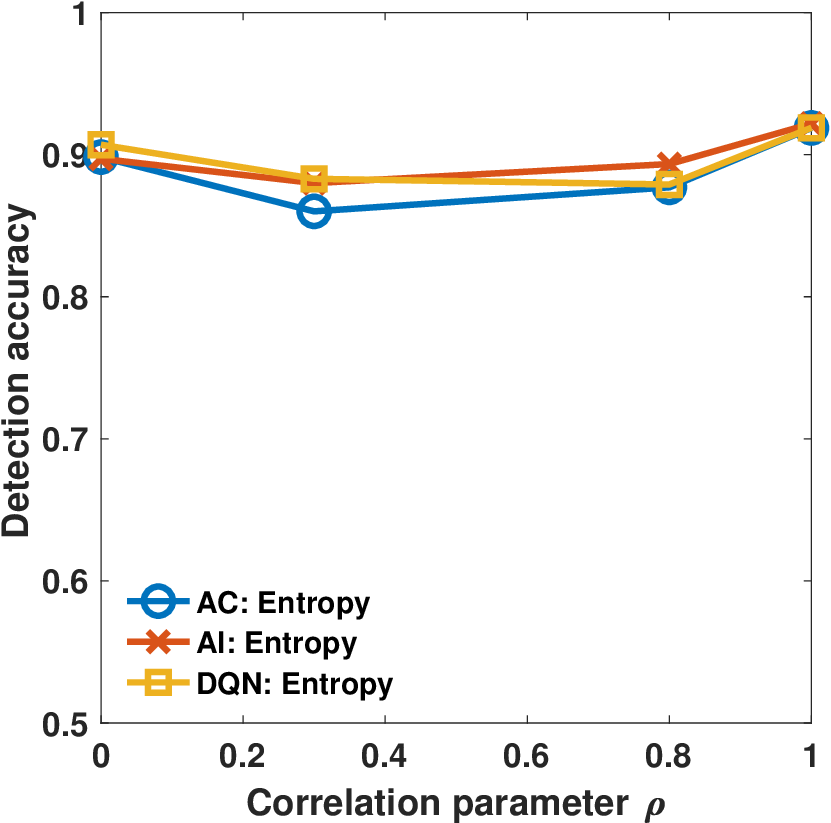}}
		\end{subfigure}
			 \hs{0.35}
		\begin{subfigure}[b]{5.1cm}
		     {\includegraphics[width=5.5cm]{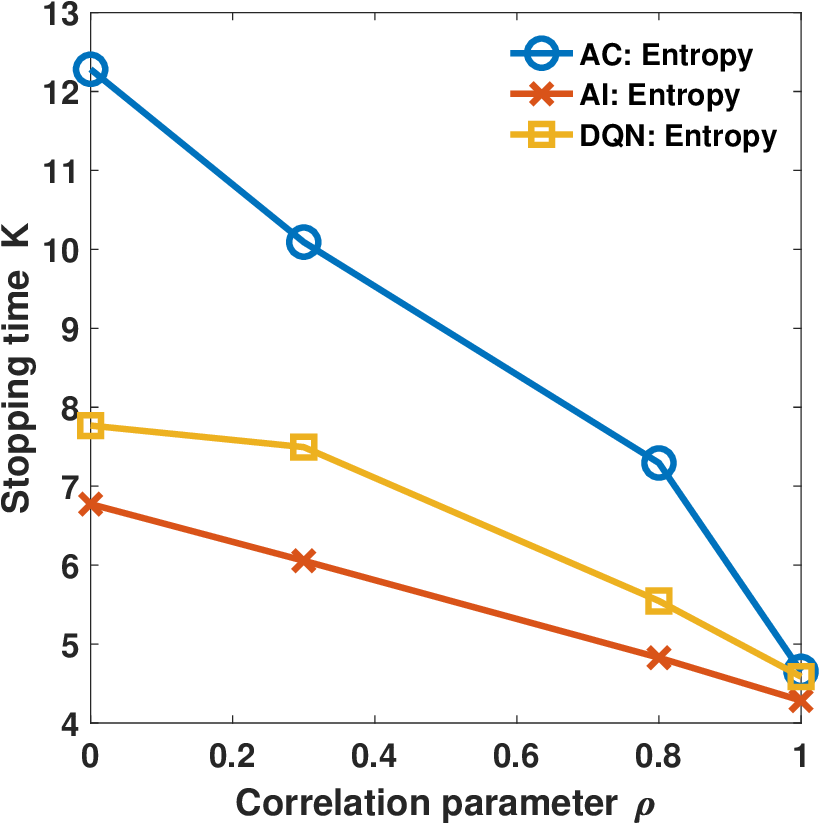}}
		\end{subfigure}
			 \hs{0.15}
		\begin{subfigure}[b]{5.1cm}
		      {\includegraphics[width=5.5cm]{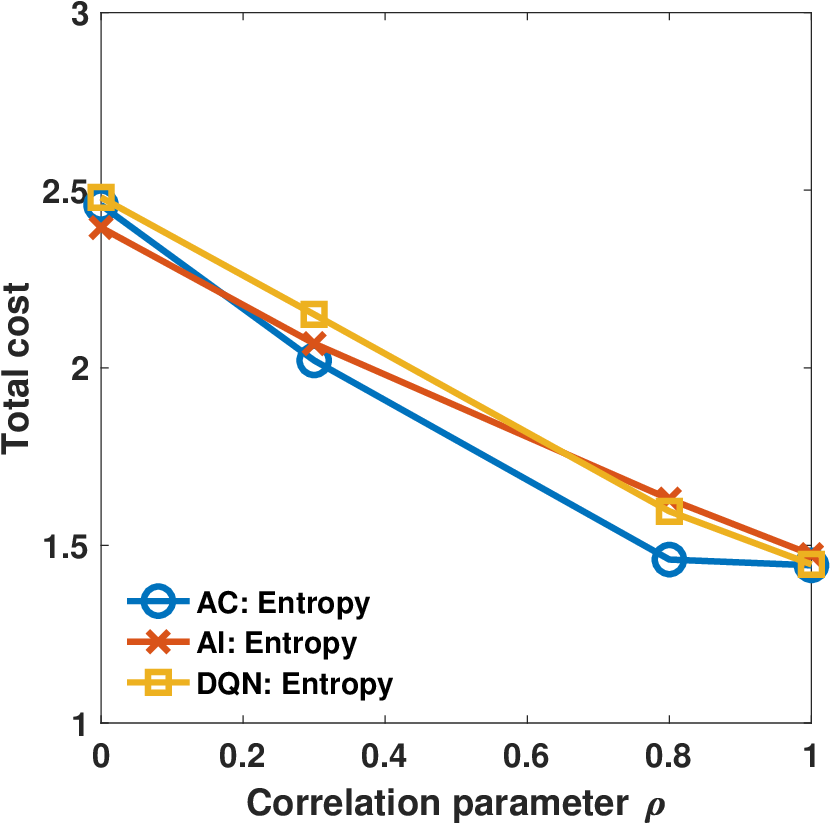}}
		\end{subfigure}
			
	\vspace{0.5cm}
	
	   \hs{-0.3}
		\begin{subfigure}[b]{5.1cm}
		{\includegraphics[width=5.7cm]{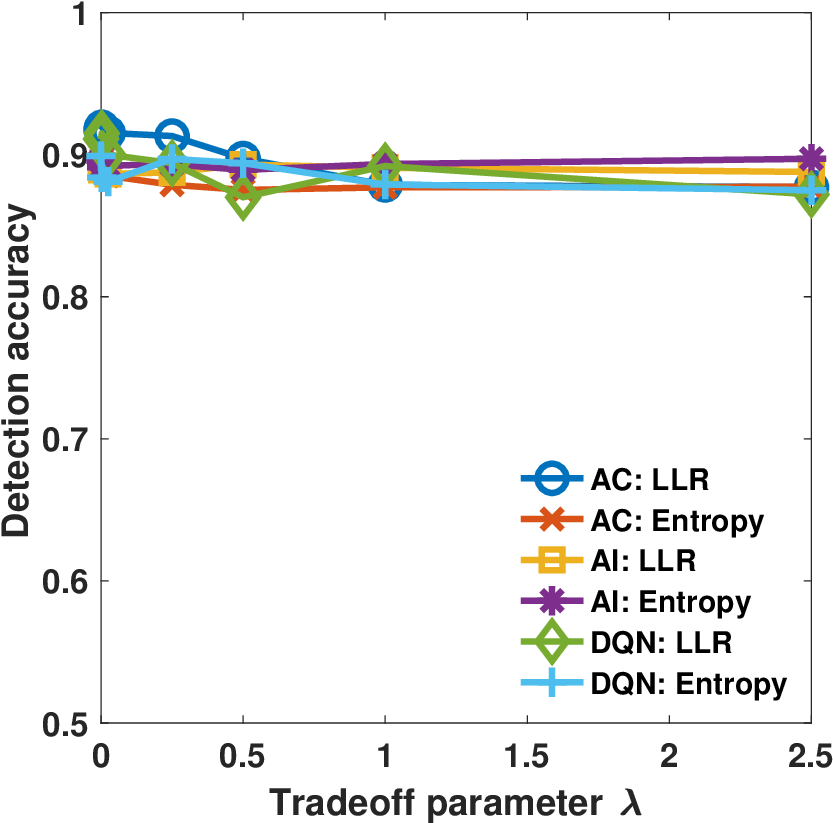}}
			\end{subfigure}
				 \hs{0.35}
		\begin{subfigure}[b]{5.1cm}
			{\includegraphics[width=5.5cm]{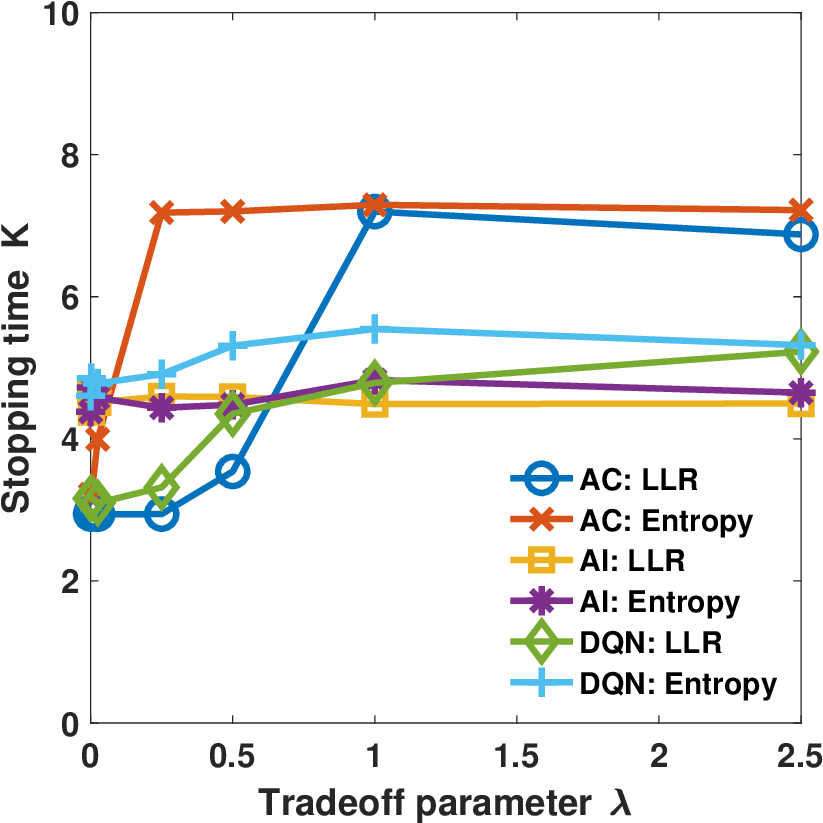}}
		\end{subfigure}
			 \hs{0.15}
		\begin{subfigure}[b]{5.1cm}
			{\includegraphics[width=5.5cm]{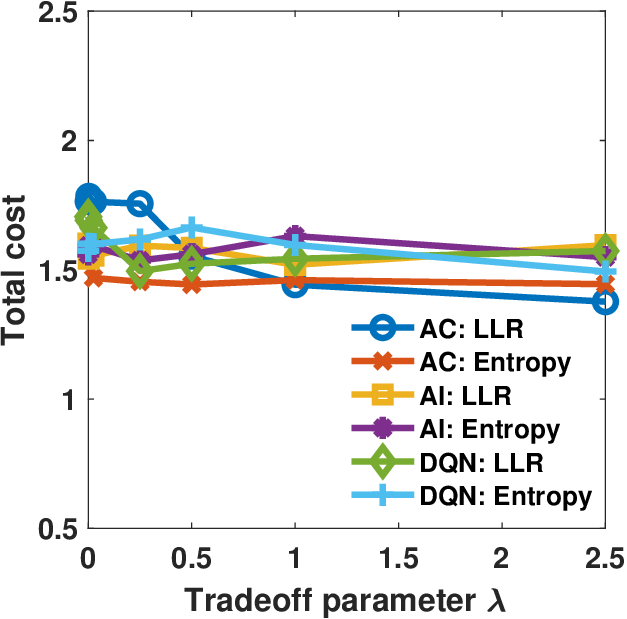}}
		\end{subfigure}		
		\end{center}
	\caption{ Performance of the actor-critic, active inference, and deep Q-learning algorithms for two different reward functions. Unless otherwise mentioned in the plot, we choose $\upi=0.8$, $\rho=0.8$, and $\lambda=1$.}
	\label{fig:parameter_vs}
\end{figure*}

\begin{figure*}[hptb]
\begin{center}
 \hs{-0.3}
		\begin{subfigure}[t]{5.3cm}
		{\raisebox{0.3cm}{\includegraphics[width=5.3cm]{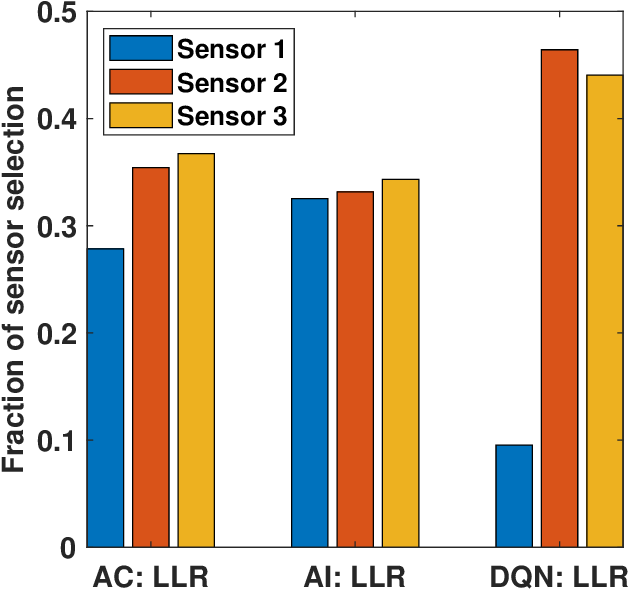}}}
			\end{subfigure}
		\begin{subfigure}[b]{5.45cm}
			{\includegraphics[width=5.45cm]{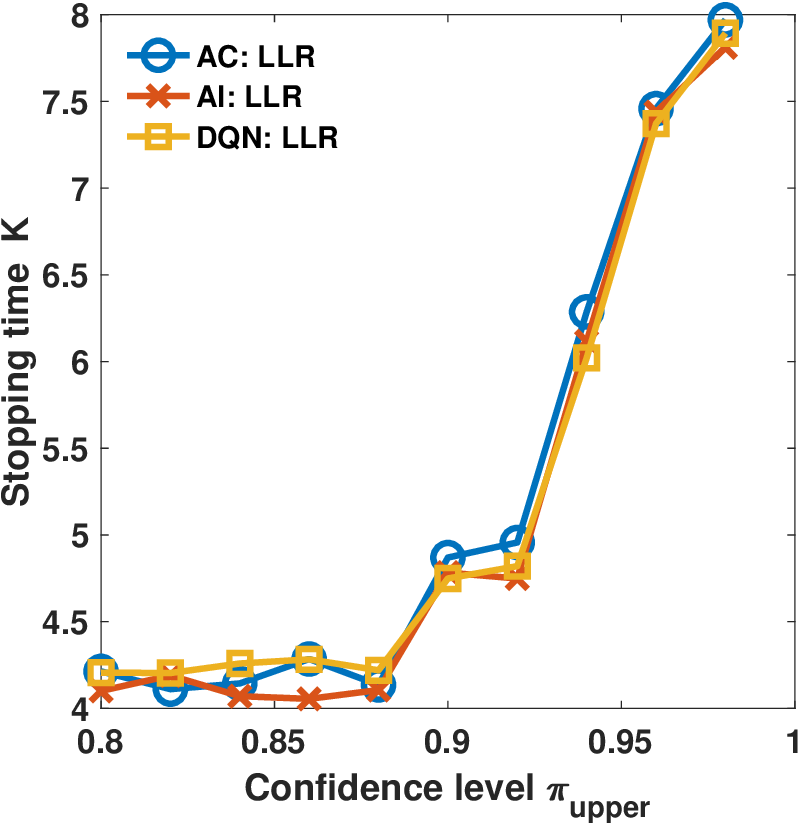}}
		\end{subfigure}
		\begin{subfigure}[b]{5.45cm}
			{\includegraphics[width=5.45cm]{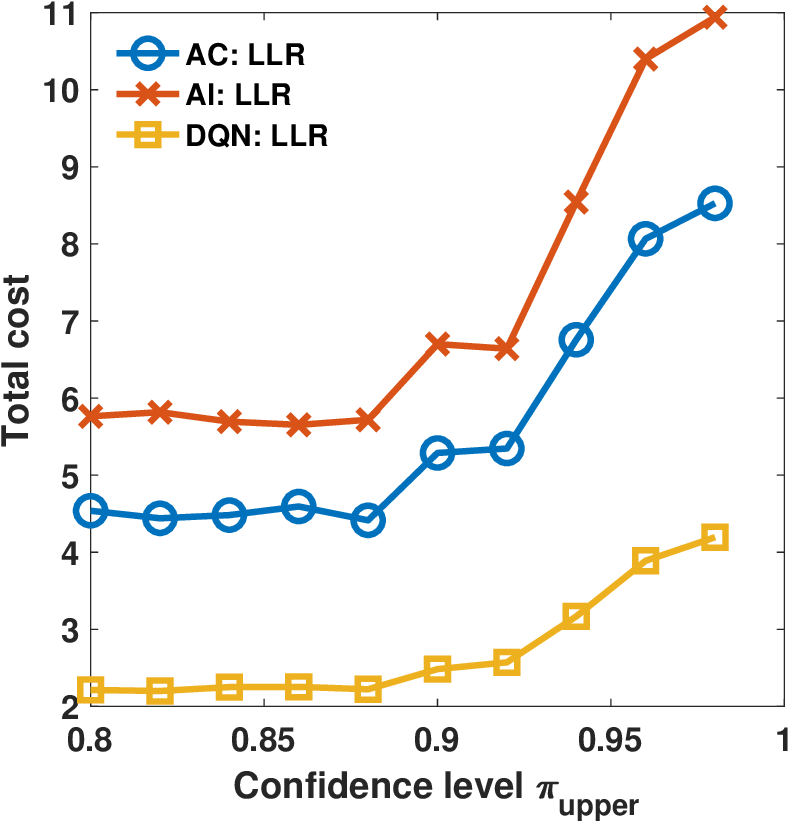}}
		\end{subfigure}		
	\end{center}
	\caption{ Performance of the actor-critic, active inference, and deep Q-learning algorithms when the sensing costs differ: $c_1=2$ and $c_2=c_3=0.2$. We choose $\rho=\lambda=1$, $p_i=0.2$ for $i=1,2,3$,  and for the bar plot, we set $\upi=0.94$. }
	\label{fig:CostDiff}
\end{figure*}

	\begin{figure*}[hptb]
\begin{center}
 \hs{-0.3}	
		\begin{subfigure}[b]{5.3cm}
		{\raisebox{0.3cm}{\includegraphics[width=5.3cm]{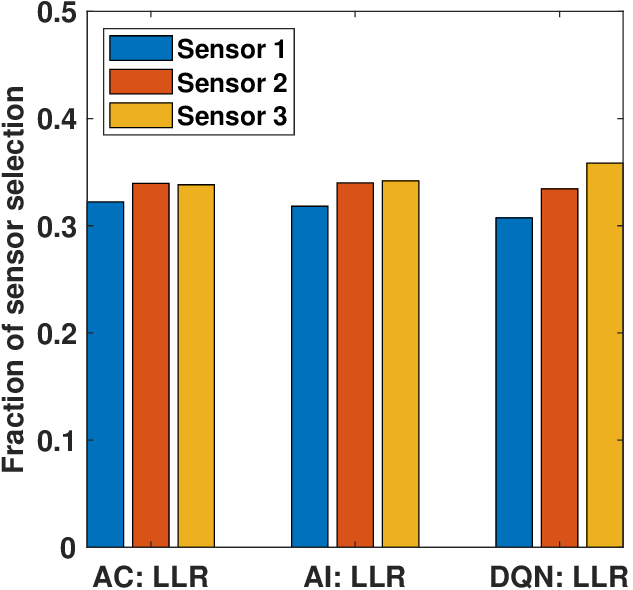}}}
			\end{subfigure}
		\begin{subfigure}[b]{5.45cm}
			{\includegraphics[width=5.45cm]{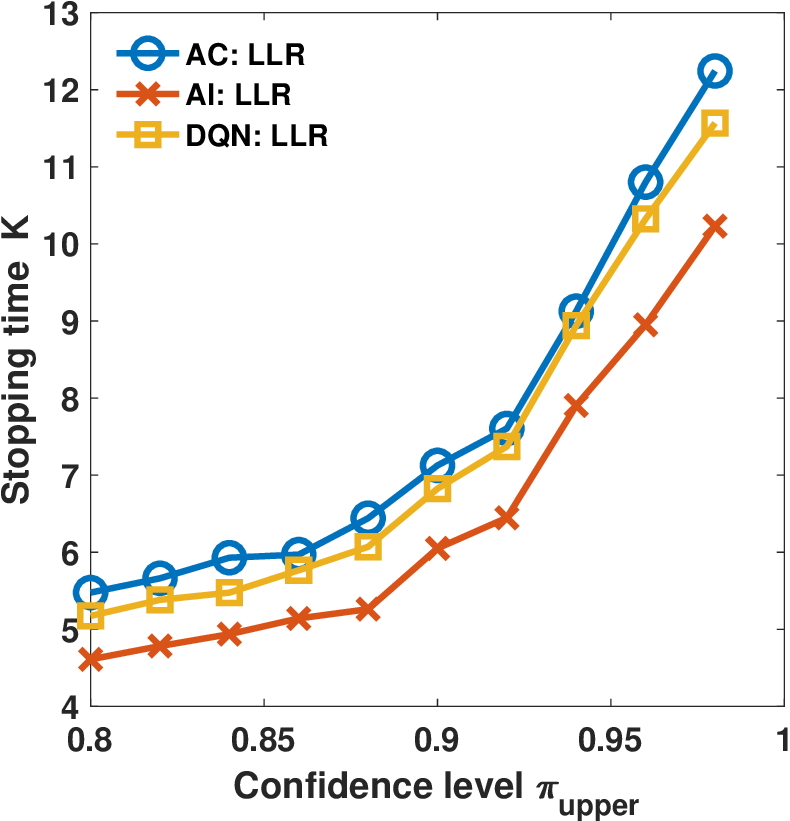}}
		\end{subfigure}
		\begin{subfigure}[b]{5.45cm}
			{\includegraphics[width=5.45cm]{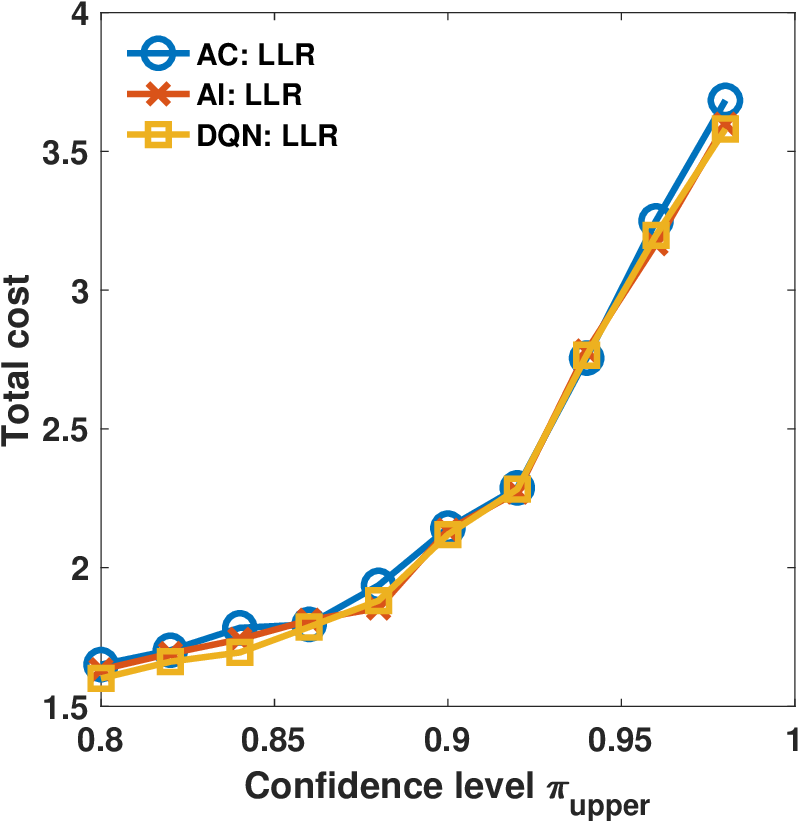}}
		\end{subfigure}		
\end{center}
	\caption{ Performance of the actor-critic, active inference, and deep Q-learning algorithms when the flipping probabilities differ: $p_1=0.45$ and $p_2=p_3=0.2$. We choose  $\rho=\lambda=1$,  $c_i=0.2$ for $i=1,2,3$,  and for the bar plot, we set $\upi=0.94$. }
	\label{fig:ProbDiff}
\end{figure*}

\begin{figure*}[hptb]
\begin{center}
 \hs{-0.3}		
		\begin{subfigure}[b]{5.3cm}
		{\raisebox{0.3cm}{\includegraphics[width=5.45cm]{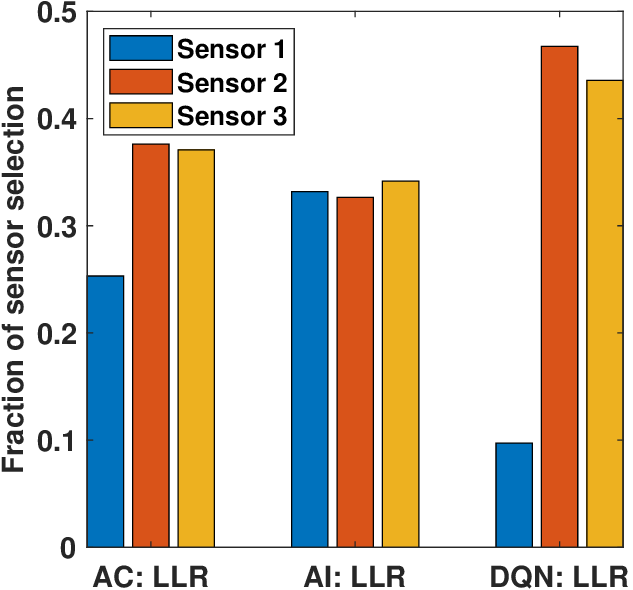}}}
			\end{subfigure}
			\hs{0.1}
		\begin{subfigure}[b]{5.45cm}
			{\includegraphics[width=5.45cm]{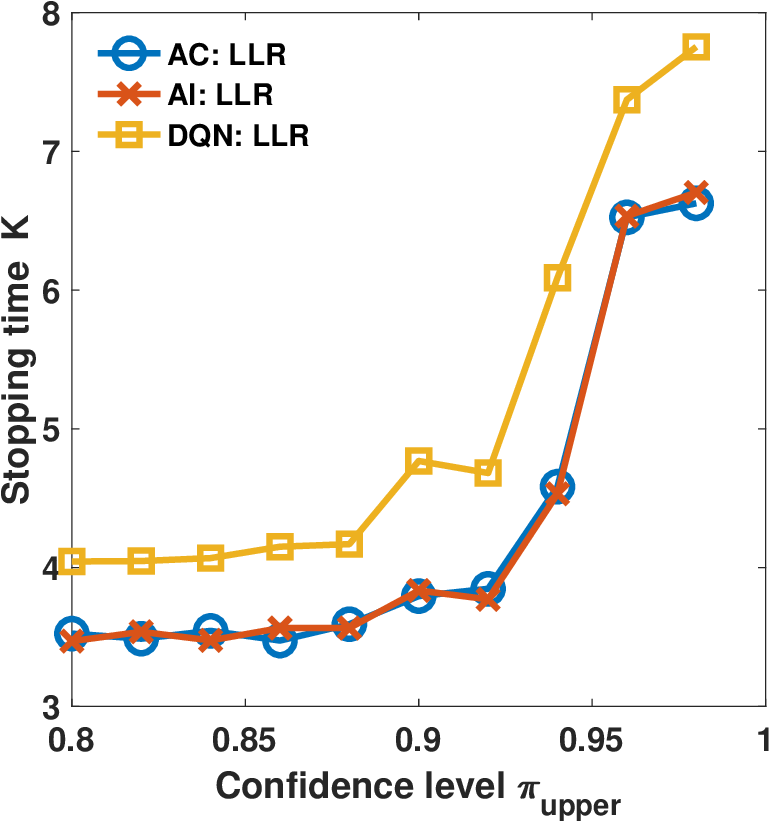}}
		\end{subfigure}
		\begin{subfigure}[b]{5.45cm}
			{\includegraphics[width=5.45cm]{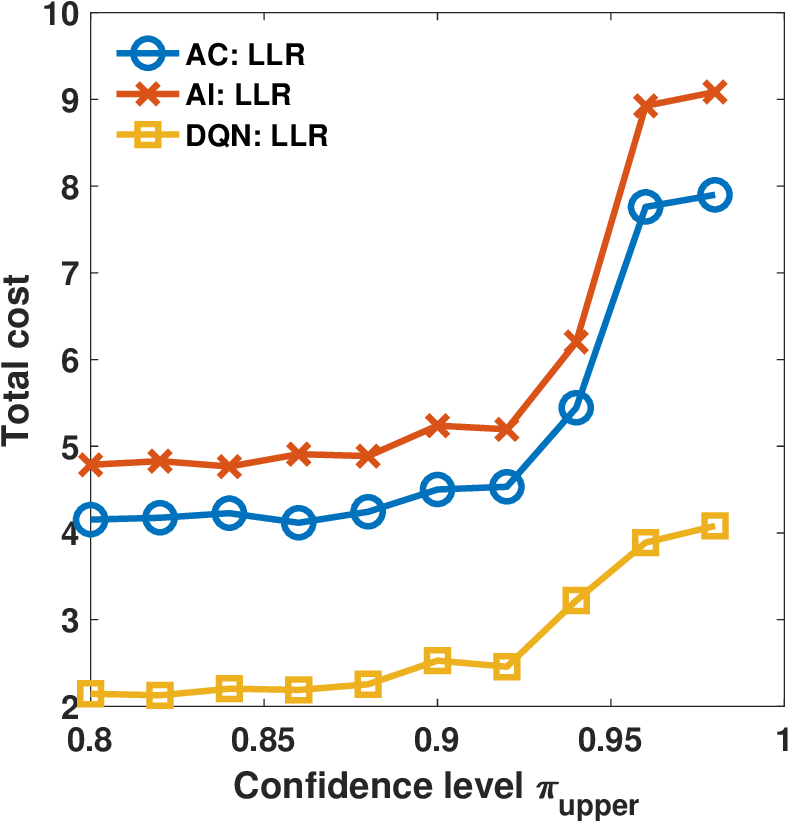}}
		\end{subfigure}		
\end{center}
	\caption{Performance of the actor-critic, active inference and deep Q-learning algorithms when both sensing costs are different: $c_1=2$ and $c_2=c_3=0.2$; and $p_1=0.02$, and $p_2=p_3=0.2$. We choose $\rho=\lambda=1$,  and we set $\upi=0.94$ for the bar plot.}
	\label{fig:AllDiff}
\end{figure*}

\begin{figure*}[hptb]
\begin{center}
 \hs{-0.3}
		\begin{subfigure}[b]{5.1cm}
		{\includegraphics[width=5.65cm]{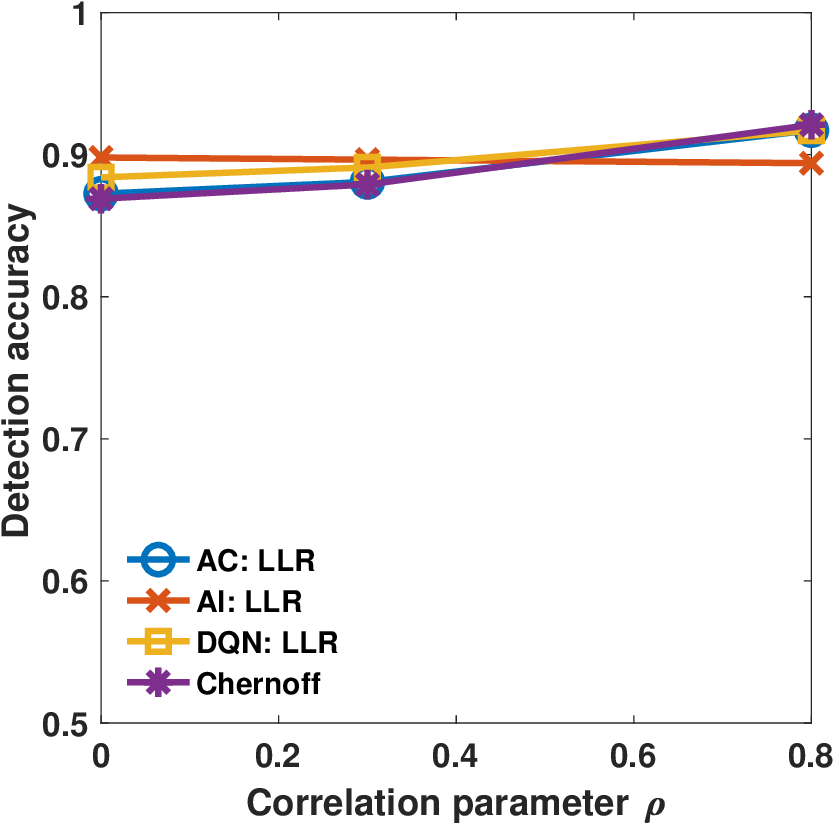}}
			\end{subfigure}
				 \hs{0.4}
		\begin{subfigure}[b]{5.1cm}
			{\includegraphics[width=5.45cm]{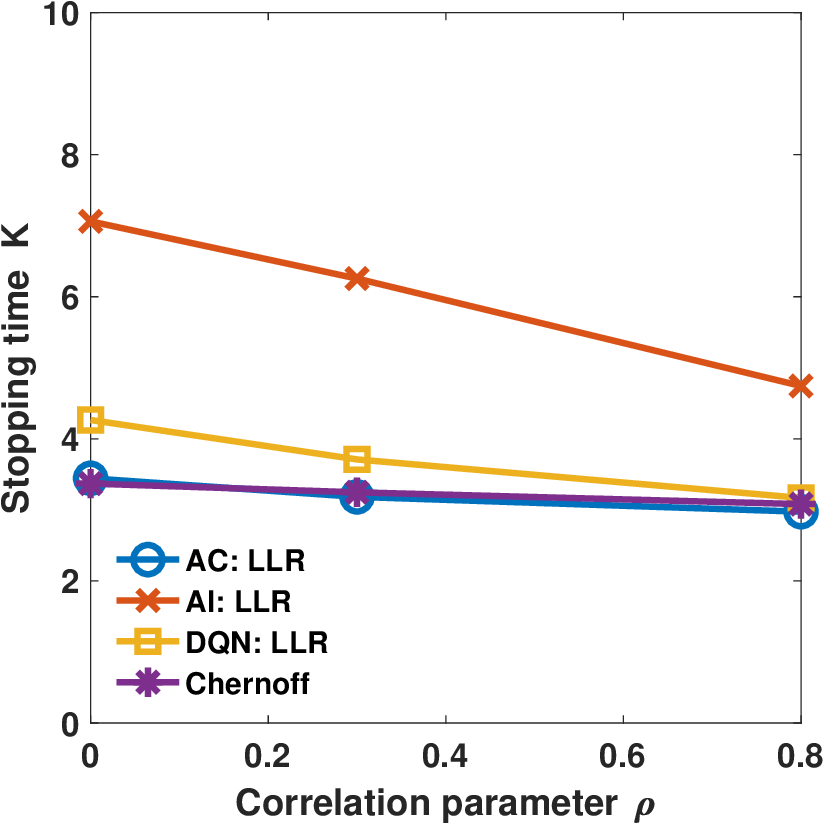}}
		\end{subfigure}
			 \hs{0.2}
		\begin{subfigure}[b]{5.1cm}
			{\includegraphics[width=5.45cm]{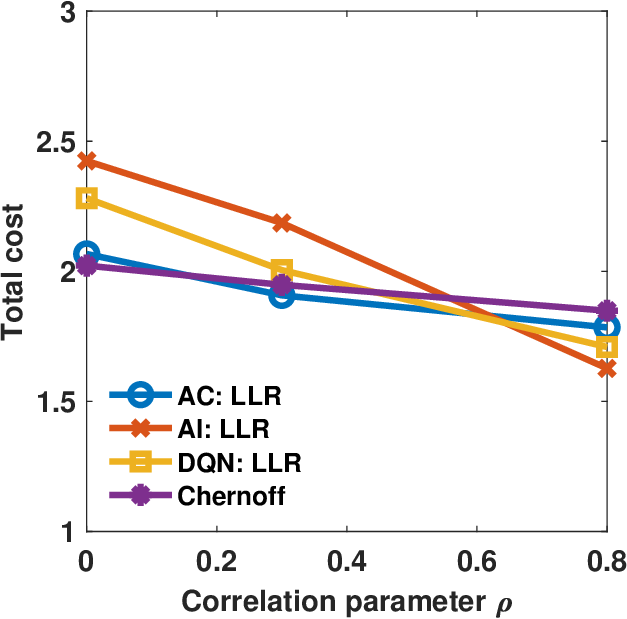}}
		\end{subfigure}		
\end{center}
	\caption{Comparison of our algorithms with the Chernoff test when $\upi=0.82$, $\lambda=0$ and $c_i=p_i=0.2$ for $i=1,2,3$.}
	\label{fig:Chernoff}
\end{figure*}

In this section, we present numerical results comparing the performances of deep RL and deep active inference algorithms. We choose the number of processes as $N=3$ and, thus, $M=2^N=8$.  The prior probability of a process being normal is taken as $q=0.8$. Here, the first and second processes are assumed to be statistically dependent, and the third is independent of the other two.  The correlation between the dependent processes is captured by the parameter~$\rho\in[0,1]$:
\begin{align*}
\bbP\lc \vecx=\begin{bmatrix}
0 & 0
\end{bmatrix}\rc &= q^2+\rho q(1-q)  \\
\bbP\lc \vecx=\begin{bmatrix}
0 & 1
\end{bmatrix}\rc =
\bbP\lc \vecx=\begin{bmatrix}
1 & 0
\end{bmatrix}\rc &= q(1-q)(1-\rho).
\end{align*}
Also, we assume that the maximum number of time slots for each episode (trial or run) is $T_{\max}=5000$.

We implement all the neural networks (the $Q$-network of deep Q-learning, the actor and critic networks, and the policy and the bootstrapped EFE networks of active inference) with three layers and the ReLU activation function between each consecutive layer. To update the network  parameters, we apply the Adam Optimizer. Also, we set $\gamma=0.9$ for the RL algorithms, and $\epsilon$ values linearly decrease from $0.4$ to $0.05$. 

{We train the neural networks over multiple episodes (realizations) where, for each episode, we choose the process states from the prior distribution mentioned above, and the number of time slots for each episode is fixed as 50. The actor-critic and active inference algorithms converge after 1000 episodes, whereas the deep Q-learning algorithm requires 2000 episodes to achieve a stable policy. So, the deep Q-learning algorithm requires more extended training than the other two algorithms.

After the training phase, we test the algorithms. We start with two illustrations in \Cref{fig:demo}. They show the realizations of the variation of the belief vector $\vecpi(k)$, sensor selection $\calA(k)$, and the corresponding observations $\vecy_{\calA(k)}(k)$ over time $k$ until the stopping time. \Cref{fig:demo_AC} shows the sensor selection of the actor-critic algorithm when the true hypothesis (process state) is $[0\;0\;1]$. Here, the posterior probability corresponding to the wrong process state $[0\;0\;0]$ was high initially due to the prior distribution. Since the true process state $[0\;0\;1]$ and the state $[0\;0\;0]$ differ only in the state of the third process, the posterior probability corresponding to the true process state $[0\;0\;1]$ is not high until the third process is observed at $k=2$. Note that at $k = 2$, the selected processes and the corresponding observations are described by $(1/0, 2/1, 3/1)$, indicating that all three processes have been chosen and the noisy observations are $[0\; 1 \; 1]$. As the probability of the true process state $[0\;0\;1]$ increases (at time $k=2$), the algorithm observes the third process more often. Finally, at time $k=6$, the  posterior probability of the true process state $[0\;0\;1]$ exceeds $\upi=0.94$ and the algorithm stops. We can make similar observations from \Cref{fig:demo_AI} where the true process state  is $[0\;0\;0]$. Due to the error in observations from the first process at $k=1$ and the second process at $k=2$, the posterior probability of the state $[1\;1\;0]$ increases initially. Then, the algorithm observes these two processes more often. This policy allows the algorithm to observe that the probability of the state $[1\;1\;0]$ decreases, and the probability of the true process state $[0\;0\;0]$ exceeds $\upi$ at $k=6$. }

{Next, we show the performance of the algorithms. Like the training phase, for each episode of the testing phase, we choose the process states from the prior distribution. The three performance metrics we use for comparison are detection accuracy, stopping time, and total cost, as defined in \Cref{sec:anomaly}. If the estimated hypothesis is the same as the true hypothesis, the (instantaneous) detection accuracy is one, and otherwise, it is zero. Also, stopping time is the shortest time at which the stopping criteria in \eqref{eq:stoppingrule} is met. The average detection accuracy, stopping time, and total cost obtained using the $10^4$ episodes are shown in Figs.~\ref{fig:parameter_vs} through \ref{fig:Chernoff}. Like the training phase, during testing for each episode, we choose the process states from the prior distribution mentioned above. In \Cref{fig:CostDiff,fig:ProbDiff,fig:AllDiff}, we also show bar plots where the heights are proportional to the fraction of times each process is chosen.} In the figures, we compare the three algorithms (label names in brackets), deep Q-learning (\texttt{DQN}), actor-critic  (\texttt{AC}), and active inference (\texttt{AI}) algorithms considering both Bayesian LLR-based (\texttt{LLR}) and entropy-based (\texttt{Entropy}) reward functions. Our observations from the numerical results are presented next.

\subsubsection{Confidence level $\upi$} The variations in the performance of different algorithms with $\upi$ are shown in the first row of \Cref{fig:parameter_vs}, and \Cref{fig:CostDiff,fig:ProbDiff,fig:AllDiff}.  All three performance metrics increase with $\upi$ in all cases. This observation is intuitive as a higher value of $\upi$ implies higher accuracy and requires the algorithms to collect more observations before they decide on anomalous processes. Also, the accuracy levels achieved by all the algorithms are comparable in all the settings because the common $\upi$ sets the desired confidence level of detection. 

\subsubsection{Correlation parameter $\rho$} The second row of \Cref{fig:parameter_vs} illustrates the  performances with varying $\rho$. The accuracy is insensitive to $\rho$ as it is decided by the confidence level $\upi$. On the other hand, the stopping time and total cost decrease with $\rho$. This decrease is expected because when the correlation increases, an observation corresponding to one of the dependent processes gives more information about the other. Consequently, the algorithms require fewer observations and a shorter stopping time to reach the same confidence level.

\subsubsection{Tradeoff parameter $\lambda$} The last row of \Cref{fig:parameter_vs} depicts the changes in the algorithm performances with $\lambda$. As in the case of $\rho$, the accuracy and total cost do not vary significantly with $\lambda$ for a fixed value of $\upi$ and $\rho$. This behavior is because when $\rho$ is fixed, we need the same number of observations to achieve the same confidence level. However, as $\lambda$ increases, each observation becomes costlier, and the stopping time increases. We notice that the stopping time of the actor-critic algorithm is more sensitive to $\lambda$ compared to the deep Q-learning and active inference algorithms. One reason for this could be that the temporal error, which is a function of only the posterior, is more sensitive to the parameter $\lambda$ than the $Q$-function learned by the deep Q-learning algorithm and EFE learned by the active inference algorithm, which are both functions of the posterior belief and action.

\subsubsection{Reward functions} From \Cref{fig:parameter_vs}, we infer that all the algorithms provide similar performance levels with both choices of the reward function. However, the actor-critic and deep Q-learning algorithms slightly underperform with the entropy-based reward function. Since the two reward functions have $\lambda\lV\calA(t)\rV$ in common, as $\lambda$ increases, the performance difference also grows, as observed from the last row of \Cref{fig:parameter_vs}. In other words, the performance gap is the largest when $\lambda=0$, and the two reward functions become identical as $\lambda$ goes to $\infty$.   Further, we recall from \Cref{sec:reward} that for the same value of $\lambda$, the Bayesian LLR reward functions give more weight to the accuracy than the cost (see \eqref{eq:compare_CH}). As a result, the performance with the entropy-based function for a particular value of $\lambda$ is similar to that with the Bayesian LLR reward for a larger value of $\lambda$.  For example, the sudden change in the stopping time of the actor-critic algorithm with $\lambda$ occurs at $\lambda=0.05$ for the entropy-based function, whereas it occurs at $\lambda=0.2$ for the Bayesian LLR-based reward. We observe similar behavior for the deep Q-learning algorithm as well.

\subsubsection{Sensing cost $c_i$ and flipping probability $p_i$} We analyze the dependence of the algorithms' performance on the sensing cost and flipping probability under three settings: 1) nonuniform sensing costs and uniform flipping probabilities (see \Cref{fig:CostDiff}); 2) uniform sensing costs and nonuniform flipping probabilities (see \Cref{fig:ProbDiff}); and 3) nonuniform sensing costs and flipping probabilities (see \Cref{fig:AllDiff}) across the processes. From \Cref{fig:CostDiff}, the deep Q-learning algorithm is more sensitive to different cost values $c_i$. In all settings considered in \Cref{fig:CostDiff}, the deep Q-learning agent chooses the first process less often, leading to the lowest total cost and best performance. The actor-critic algorithm also adapts to the varying cost, while the active inference algorithm is relatively less insensitive to the different costs. Similarly, when we increase the flipping probability of the first process in \Cref{fig:ProbDiff} (with uniform sensing costs), we see that all algorithms adapt their policies. However, the policy offered by the active inference algorithm has shorter stopping times than the other algorithms for comparable values of the total cost. The differences in the policies of the three algorithms are more evident in \Cref{fig:AllDiff} when we vary both sensing cost and flipping probabilities. In this setting, the deep Q-learning algorithm chooses the first process less often despite its smaller flipping probability. As in the case of \Cref{fig:ProbDiff}, active inference is more sensitive to the flipping probability than the cost, and as a result, it gives the shortest stopping times at the price of a higher total cost. The performance of the actor-critic algorithm is between those of the other two algorithms. The actor-critic algorithm provides stopping times comparable to those of the active inference algorithm while incurring a smaller total sensing cost.

\subsubsection{Competing algorithms}
\label{sec:recommendation}
{We first note that all the algorithms have similar detection accuracy due to the common stopping criteria in \eqref{eq:stoppingrule}, i.e., they stop only when the detection accuracy of the algorithm always exceeds $\upi$. So, the choice of the best learning algorithm depends on the stopping time and total cost. We first look at the algorithm performances for the uniform cost and filliping probability case from \Cref{fig:parameter_vs,fig:Chernoff}. For small values of $\lambda$, the actor-critic algorithm offers the best stopping time but has a slightly higher cost than the other algorithms. As $\lambda$ increases, its stopping time also increases, and the active inference algorithm provides the best stopping time for a comparable total cost.  Also, the active inference algorithm offers slightly better performance than deep Q-learning. However, our experiments show that the Q-learning algorithm requires more episodes in the training phase than the other algorithms to achieve a stable policy. The memory replay in the Q-learning algorithm also makes its training phase further longer than the other algorithms. Therefore, the actor-critic algorithm is more suitable for sensing cost-critical applications, and for time-sensitive applications, we recommend the active inference algorithm over Q-learning. 
 
 Next, we look at the nonuniform setting in \Cref{fig:CostDiff,fig:ProbDiff,fig:AllDiff}. We notice that the deep Q-learning algorithm is more sensitive to the nonuniform sensing cost, whereas the active inference algorithm is more sensitive to the nonuniform flipping probability. So, in the nonuniform setting, we prefer Q-learning for cost-critical applications and  
active inference for stopping time-sensitive applications. These observations further justify our joint analysis of different learning-based methods.} 

\subsubsection{Comparison with Chernoff test}  \Cref{fig:Chernoff} compares our algorithms with the classical Chernoff test. The stopping time and the total sensing cost of the Chernoff test are relatively insensitive to the variation in $\rho$. In contrast, our algorithms, particularly the active inference algorithm, adapt their stopping time and total sensing cost to $\rho$. This observation is intuitive as the policy followed by the Chernoff test does not depend on $\rho$ or $\lambda$, and it assumes that the processes are independent. Therefore, \eqref{eq:chernoff} leads to the optimum performance when $\rho=0$ but deteriorates as $\rho$ increases.


\section{Conclusion}\label{sec:con}

This paper considered the anomaly detection problem, where the goal is to identify the anomalies among a given set of processes. We modeled the problem of anomaly detection as an MDP problem aiming at the detection accuracy exceeding a desired value while minimizing the delay and total sensing cost. To this end, we designed two objective functions based on Bayesian LLR and entropy and presented two deep RL-based algorithms and a deep active inference algorithm. Through simulation results, we compared our algorithms and showed that all algorithms perform similarly in the detection accuracy for the same confidence level. However, the dueling deep Q-learning algorithm required a more prolonged training phase, and the active inference algorithm is more robust to the trade-off parameter and adapts better to the correlation parameter.  We also inferred that the policy of the dueling deep Q-learning algorithm always led to more negligible sensing costs. In contrast, the active inference algorithm is more sensitive to the flipping probabilities. Extending our algorithm to track any changes in the behavior of the processes over a more extended time period is an exciting future direction.

\bibliographystyle{IEEEtran}
\bibliography{AnomalyDetection}

\end{document}